\documentclass[lettersize,journal]{IEEEtran}

\usepackage{amsmath}
\usepackage{amsthm}
\usepackage{epstopdf}
\usepackage{extarrows}
\usepackage{amsfonts}
\usepackage{amssymb}
\usepackage{verbatim}
\usepackage{multirow}
\usepackage{graphicx}
\usepackage{subfigure}
\usepackage{graphicx}
\usepackage{algorithmic}
\usepackage{color}
\usepackage{cite}
\usepackage{hyperref}
\usepackage{epstopdf}
\usepackage[linesnumbered,lined,ruled]{algorithm2e}
\usepackage{booktabs}
\usepackage{threeparttable}
\usepackage{rotating}
\usepackage{footmisc}
\usepackage{float}
\usepackage{makecell}
\usepackage{float}
\usepackage[switch]{lineno}

\definecolor{purple}{RGB}{127,0,127}    % 3.RGB

\begin{document}
\title{Hierarchical Multi-Relational Graph Representation Learning for Large-Scale Prediction of Drug-Drug Interactions}

\author{Mengying~Jiang,
        Guizhong Liu,~\IEEEmembership{Member,~IEEE,} 
        Yuanchao~Su,~\IEEEmembership{Senior Member,~IEEE,} \\
        Weiqiang Jin,
        Biao Zhao
             
\thanks{This study was supported in part by the Shaanxi Key Research and Development Program under Grant 2018ZDCXL-GY-04-03-02, in part by the National Natural Science Foundation of China under Grant 42001319, and in part by the Scientific Research Program of the Education Department of Shaanxi Province under Grant 21JK0762.(\textit{Corresponding author: Guizhong Liu}.)}
\thanks{M. Jiang, G. Liu, W. Jin, and B. Zhao are with the School of Electronics and Information Engineering, Xi'an Jiaotong University, Xi'an 710049, China (e-mail: myjiang@stu.xjtu.edu.cn, liugz@xjtu.edu.cn, weiqiangjin@stu.xjtu.edu.cn, biaozhao@xjtu.edu.cn).}
\thanks{Y. Su is with the College of Geomatics, Xi'an University of Science and Technology, Xi'an, 710054, China; is also with Key Laboratory of Computational Optical Imaging Technology, Aerospace Information Research Institute, Chinese Academy of Sciences, Beijing, 100094, China (e-mail: suych3@xust.edu.cn).}
} 

\markboth{Journal of \LaTeX\ Class Files,~Vol.~14, No.~8, August~2021}%
{Shell \MakeLowercase{\textit{et al.}}: A Sample Article Using IEEEtran.cls for IEEE Journals}

% Remember, if you use this you must call \IEEEpubidadjcol in the second column for its text to clear the IEEEpubid mark.

\maketitle
% \linenumbers

\begin{abstract}
\label{abs}
Most existing methods for predicting drug-drug interactions (DDI) predominantly concentrate on capturing the explicit relationships among drugs, overlooking the valuable implicit correlations present between drug pairs (DPs), which leads to weak predictions.
To address this issue, this paper introduces a hierarchical multi-relational graph representation learning (HMGRL) approach.
Within the framework of HMGRL, we leverage a wealth of drug-related heterogeneous data sources to construct heterogeneous graphs, where nodes represent drugs and edges denote clear and various associations. 
The relational graph convolutional network (RGCN) is employed to capture diverse explicit relationships between drugs from these heterogeneous graphs. 
Additionally, a multi-view differentiable spectral clustering (MVDSC) module is developed to capture multiple valuable implicit correlations between DPs. 
Within the MVDSC, we utilize multiple DP features to construct graphs, where nodes represent DPs and edges denote different implicit correlations.
Subsequently, multiple DP representations are generated through graph cutting, each emphasizing distinct implicit correlations. 
The graph-cutting strategy enables our HMGRL to identify strongly connected communities of graphs, thereby reducing the fusion of irrelevant features.
By combining every representation view of a DP, we create high-level DP representations for predicting DDIs. 
Two genuine datasets spanning three distinct tasks are adopted to gauge the efficacy of our HMGRL.
Experimental outcomes unequivocally indicate that HMGRL surpasses several leading-edge methods in performance.
Our source codes are available at the following link \href{https://github.com/mengyingjiang/HMGRL}{https://github.com/mengyingjiang/HMGRL.} 

\end{abstract}
\begin{IEEEkeywords}
Graph Representation Learning, Large-Scale Predictor, Heterogeneous Data Sources, Spectral Clustering.
\end{IEEEkeywords}

\IEEEpeerreviewmaketitle

\section{Introduction} \label{Introduction}
\IEEEPARstart{P}harmaceutical drugs are chemical substances or compounds designed to diagnose, prevent, and cure diseases~\cite{IIFDTI}. Coadministration can enhance efficacy and mitigate drug resistance, emerging as a promising approach in drug therapy~\cite{RANEDDI}. Nevertheless, drugs may interact when administered concurrently, resulting in unexpected drug-drug interactions (DDIs) that can cause adverse reactions, including allergies and even death~\cite{SGRL-DDI}. Consequently, accurately identifying DDIs is paramount for drug discovery and safe coadministration~\cite{MADRL}.
Traditional methods mainly rely on clinical trials and post-marketing surveillance to detect DDIs. However, these methods are generally time-consuming and labor-intensive, highlighting the need for more efficient alternatives~\cite{DDIMDL}.
Recently, computational approaches have gained popularity among researchers due to low cost and high accuracy~\cite{MADRL}.

In contemporary times, many large-scale comprehensive knowledge bases, including DrugBank~\footnote{\url{https://go.drugbank.com}}, 
PubChem~\footnote{\url{https://pubchem.ncbi.nlm.nih.gov/}},
ChEMBL~\footnote{\url{https://www.surechembl.org/search/}}, 
and PDB~\footnote{\url{https://www.rcsb.org/}}, provide abundant heterogeneous drug data sources that bolster computational approaches.
Such wealth of information empowers computational approaches to yield more precise predictions, thereby rendering large-scale DDI prediction viable.
Computational approaches can broadly be classified into binary and multi-relational DDI prediction~\cite{GoGNN}.
Binary DDI prediction focuses on determining the existence of interactions and can be formulated as a binary classification task~\cite{AMDE}. Multi-relational DDI prediction aims to identify the specific type of DDIs and is formulated as a multi-classification task~\cite{zhang2015label}.
Generally, multi-relational DDI prediction offers more detailed information on DDIs regarding pharmacological effects, which are valuable for investigating potential causal mechanisms underlying DDI occurrences~\cite{MDF-SA-DDI}.

Recently, many multi-relational DDI prediction methods have been proposed~\cite{MP1}. 
Deng et al. \cite{DDIMDL} utilized multiple biochemical attributes, including targets, molecular substructures, pathways, and enzymes, to compute drug similarities and built a deep neural network (DNN) model for DDI prediction. Lee et al.~\cite{lee} also used multiple drug similarities as model inputs and employed autoencoders to learn drug embeddings for DDI prediction.
Zhang et al.~\cite{zhang2015label} constructed a network based on drugs' side effects and structural similarities, then identified DDIs using a label propagation algorithm.
Yu et al.~\cite{RANEDDI} built the multi-relational DDI graph to obtain the relation-aware network structural information of drugs for improving DDI prediction.
Hong et al.~\cite{LaGAT} derived drug embeddings by aggregating features of neighboring drugs from different attention pathways, with the attention weights relying on the feature representations of the drugs and their neighbors.
Although the adept use of prior knowledge endows these methods to deal with DDI prediction tasks, they frequently neglect the intrinsic associations between DPs. This oversight may limit the potential of these techniques, suggesting opportunities for further refinement and enhancement.

% Additionally, more researchers have begun to focus on correlations between entity pairs and developed some related works \cite{DeepPSE,IIFDTI}.
% Zhao et al.~\cite{GCN-DTI} proposed a drug-protein interaction prediction method that constructs a drug-protein pair graph where nodes represent drug-protein pairs, and edges indicate correlations between drug-protein pairs. Specifically, the correlation is influenced by whether two drug-protein pairs share a common drug or protein. Each drug-protein pair can aggregate features from connected drug-protein-pairs through the constructed graph.
% Lin et al.~\cite{MDF-SA-DDI} employed multiple drug-related attributes (such as molecular substructure, targets, and enzyme) to compute various similarities for representing drugs and subsequently used transformer layers to uncover the underlying associations between DPs for performing DP feature fusion.
% These two methods can capture correlations between entity pairs and have achieved remarkable prediction performance in their respective tasks.
% However, most existing methods do not fully exploit the various correlations between DPs, leaving substantial room for improvement.

Drug molecules and DDI data are inherently graph-structured. Graph neural networks (GNNs) harness neural networks to distill features from such data, standing out as the premier technique for graph representation learning\cite{bigd1,bigd4}. In the past few years, a surge of artificial neural network designs rooted in GNNs has emerged\cite{bigd2}. Notable ones include the graph convolutional network (GCN)\cite{GCN+chapter2017}, graph attention network (GAT)\cite{GAT+chapter2018}, graph isomorphism network (GIN)\cite{GIN+chapter2019}, and GraphSAGE\cite{GraphSage+chapter2017}.
For GNNs, each node iteratively updates and improves its embeddings by aggregating the features of its neighbors and the node itself~\cite{ASTF}. 
Recently, many researchers have introduced the GNN architecture for designing DDI prediction models. Examples include the link-aware graph attention network (LaGAT)\cite{LaGAT}, the graph of graphs neural network (GoGNN)\cite{GoGNN}, attention-mechanism-based multidimensional feature encoder (AMDE)\cite{AMDE}, contrastive self-supervised graph neural network (CSGNN)\cite{CSGNN}, and social theory-enhanced graph representation learning (SGRL-DDI)~\cite{SGRL-DDI}.
Nevertheless, most GNN-rooted methods zero in on explicit relationships, such as atomic bonds or interactions and remedies between drugs and diseases. They often overlook certain nuanced implicit correlations that remain elusive for GNNs to discern and harness effectively.

Spectral clustering (SC), derived from graph theory, emerges as a novel clustering methodology\cite{bigd3,mdsc}. Unlike most conventional clustering algorithms, SC is not confined to data adhering to a Gaussian distribution \cite{SC01}. Instead, SC forms undirected graphs from data, capturing intricate data structures, and subsequently embeds the input data into a reduced-dimensional space through graph cutting \cite{SC02}. In recent years, SC has found successful applications across diverse domains, including gene expression analysis, information processing, social network partitioning, and image segmentation, to name a few \cite{SC0}.
A surge of interest has also been observed in SC-inspired drug-drug interaction (DDI) prediction methods. Notable examples include the signed graph filtering-based neural networks (SGFNNs) \cite{SGFNNs}, skip-graph neural networks (SkipGNN) \cite{SkipGNN}, and global confounding concept discovery (GCCD) \cite{GCCD}. However, a common limitation among these methods is their reliance on traditional SC algorithms. They derive clustering results via eigendecomposition, a process that is both computationally intensive and non-differentiable.

Considering the abovementioned issues, capturing both explicit relationships between drugs and valuable implicit associations between DPs is vital for accurate DDI identification. 
Addressing this, we propose a novel method, the hierarchical multi-relational graph representation learning, abbreviated HMGRL.
Our HMGRL contains three pivotal stages: relation-aware graph structure embedding (RaGSE), multi-source feature learning, and multi-view differentiable spectral clustering (MVDSC).
In the RaGSE stage, we collect abundant drug-related data from DrugBank.
These data are used to construct two heterogeneous graphs with explicit physical connections, a multi-relational DDI graph, and a multi-attributes-based drug-drug-similarity (DDS) graph.
We employ the RGCN model\cite{rgcn} to capture drugs' RaGSEs by aggregating neighbors' features under different relations within the DDI and DDS graphs.
Afterward, a multi-source feature learning module is designed.
To derive comprehensive DP features, this module integrates multiple drug data sources, including RaGSEs, targets, enzymes, molecular substructures, and Simplified-Molecular-Input-Line-Entry-System (SMILES) strings.
Subsequently, inspired by spectral clustering (SC), this paper introduces an MVDSC module to capture multiple valuable implicit correlations between DPs.
MVDSC uses multiple DP features to establish graphs, wherein nodes symbolize DPs and edges reflect distinct implicit connections.
Multiple DP representations are yielded through graph cutting, and each highlights unique intrinsic correlations.
Amalgamating these multifaceted representation views furnishes us with a sophisticated DP depiction, laying a sturdy groundwork for DDI predictions. 
Overall, rich drug data, knowledge, and a well-designed network structure enable our HMGRL to achieve accurate predictions for DDIs. 

The main contributions of our work can be summarized as follows: 

\begin{itemize}
\item[] 
1) The proposed HMGRL can hierarchically capture diverse explicit relationships among drugs and multifaceted implicit associations between DPs.
\end{itemize}

\begin{itemize}
\item[] 
2) An innovative multi-source feature learning module is developed, integrating abstract drug embeddings with original biochemical attributes. This integration ensures a richer capture of DP features and facilitates gradient propagation.
\end{itemize}

\begin{itemize}
\item[] 
3) A novel MVDSC module is presented, allowing HMGRL to gain insights into the implicit correlations between DPs from diverse angles.
\end{itemize}

% The rest of the paper is organized as follows. 
% Section~\ref{related work} reviews previous studies relevant to this work.
% Section~\ref{Method} comprehensively describes the proposed model, HMGRL.
% Section~\ref{experiment} presents the parameter analysis, ablation studies, and experimental comparison of HMGRL with other baseline methods, followed by a case study to validate the effectiveness of HMGRL.
% Section~\ref{Conclusion} summarizes the paper.

\section{Related Work}
\label{related work}

\subsection{Relational Graph Convolutional Network}
\label{RGCN}

Let a multi-relational graph $\mathcal{G}=(\mathcal{V}, \mathcal{E}, \mathcal{R})$, where $\mathcal{V}$ represents the set of nodes, $\mathcal{E}$ denotes the set of edges, and $\mathcal{R}$ represents the set of edge relations.
$N=|\mathcal{V}|$ is the number of nodes, and $R=|\mathcal{R}|$ denotes the number of relation types.
Let ${\bf X}= \left\{{\bf x}_v\right\}_{v=1}^N \in \mathbb{R}^{N \times d}$ represents features for all nodes, and $d$ means the dimension of node features.
$\mathcal{E}=\left\{\mathcal{E}_r\right\}_{r=1}^R$, and $\mathcal{E}_r$ represents the set of edges under relation $r$. 
Let $\mathcal{A}=\left\{{\bf A}_r\right\}_{r=1}^R\in \mathbb{R}^{N \times N \times R}$ be the multi-relational adjacency tensor, where ${\bf A}_r\in \mathbb{R}^{N \times N}$ represents the adjacency matrix under relation $r$.
Let $A_{r(u,v)}$ $(u,v = 1,\dots,N)$ be the element of $\bf{A}_{r}$,
$A_{r(u,v)}=1$ if $(u,v) \in \mathcal{E}_{r}$ and $A_{r(u,v)}=0$ if $(u,v) \notin \mathcal{E}_{r}$.

The node features can be updated by aggregating their neighbor’s features under different relations \cite{rgcn}.  
The forward propagation function is defined in the following way:
\begin{equation}
{\bf \overline x}_v=\sigma\left(\sum_{r \in \mathcal{R}} \sum_{u \in \mathcal{N}_v^r} \frac{\hat{A}_{r(u,v)}}{R_v} \mathbf{x}_u{\bf W}_r + {\bf x}_v\mathbf{W}_o\right),\\
\label{rgcn}
\end{equation}
where ${\bf \overline x}_v\in \mathbb{R}^{1 \times d'}$ represents the embeddings of node $v$.
The relation-specific weight matrix, $\mathbf{W}_r \in \mathbb{R}^{d \times d'}$, supports multiple edge types.
$\mathbf{W}_0\in \mathbb{R}^{d \times d'}$ is a single weight matrix regardless of relations.
$\mathcal{N}_{v}^r$ represents the set of nodes connected to node $v$ under relation $r$.
$\sigma(\cdot)$ is an element-wise activation function: $\operatorname{ReLU}(\cdot)=\max (0, \cdot)$.
To normalize the incoming messages of each node, a set of normalized constants $\left\{R_{v}\right\}_{v=1}^N$ is calculated. 
Specifically, $R_v$ equals the number of relation types in which node $v$ is involved. 
$\hat{A}_{r(u,v)}$ is the element in $\hat{\bf A}_{r}$, and means the edge weight between nodes $u$ and $v$.
$\hat{\bf A}_{r}$ is calculated by on a classic graph-based normalization method \cite{GCN} as follows:
\begin{equation}
\hat{\bf A}_{r} = {\bf D}_{r}^{-1/2}{\bf A}_{r}{\bf D}_{r}^{-1/2},
\label{norm}
\end{equation}
where ${\bf D}_r = \operatorname{diag}\left( {\bf A}_r {\bf 1}_N\right)$ is the degree matrix of ${\bf A}_r$. 

% \begin{figure}[t]
% \centering
% \includegraphics[width=3.5in,height=1.4in]{Mainpic/3tasks.pdf}
% \vspace{-0.4cm}
% \caption{Examples of construction strategies for three different test sets. In these graphs, nodes represent drugs, solid edges indicate known DDI interactions (training set), and edges with different colors denote various interaction types. The dotted edges represent the prediction task (test set).  
% }
% \vspace{-0.3cm}
% \label{3tasks}
% \end{figure}

\begin{figure*}[t]
\centering
\includegraphics[width=7.1in,height=3.75in]{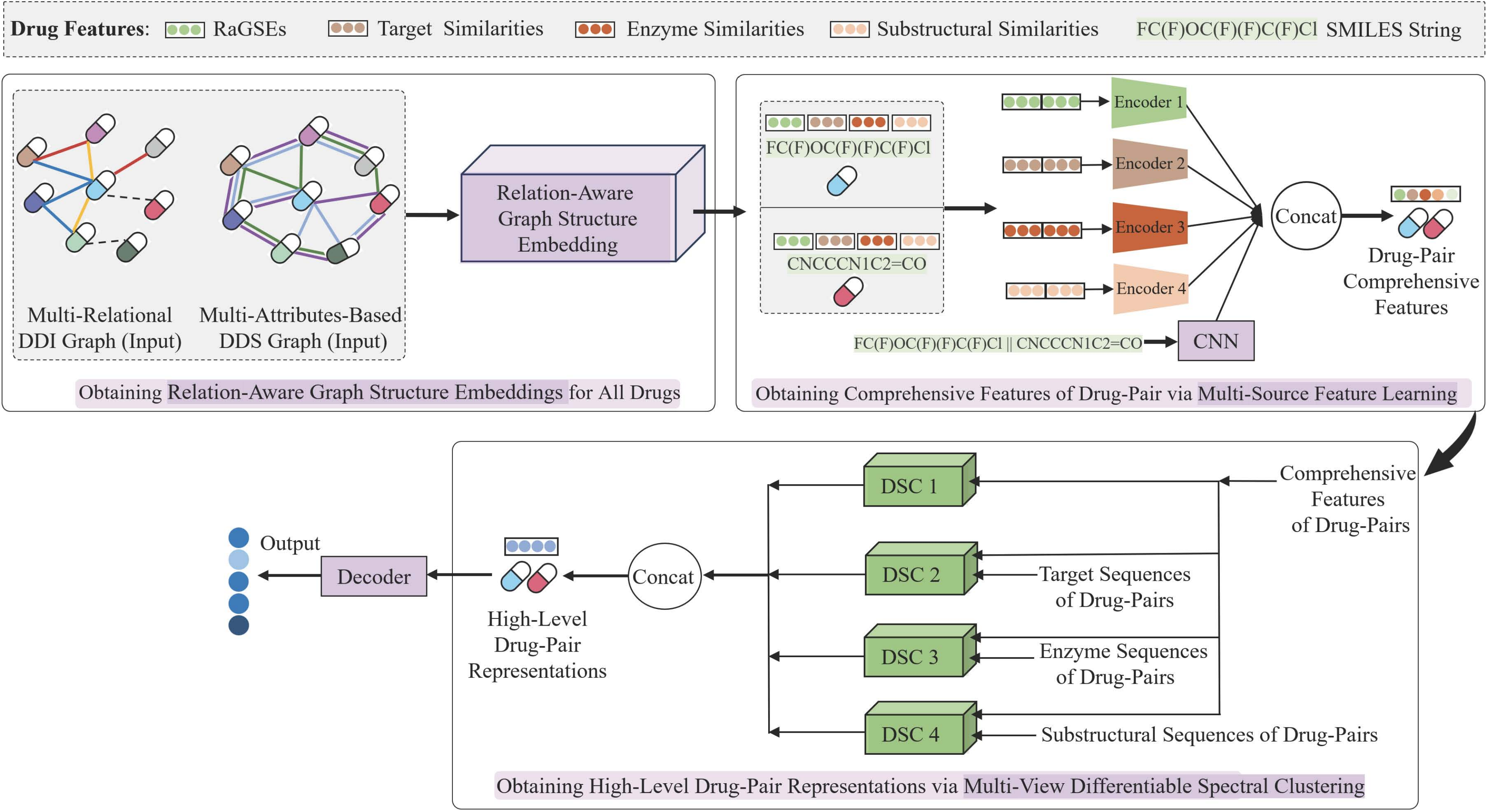}
\vspace{-0.4cm}
\caption{Flowchart of the proposed HMGRL. Firstly, HMGRL utilizes DDI and DDS graphs to produce RaGSEs for all drugs. Secondly, within the multi-source feature learning module, HMGRL integrates RaGSEs with multiple biochemical attributes to generate comprehensive DP features.
Thirdly, HMGRL develops an MVDSC module that employs multiple DP features to uncover implicit correlations between DPs.
The MVDSC module generates multiple DP representations, each highlighting distinct implicit correlations.
Ultimately, by merging all these DP representation views, we obtain a high-level DP representation suitable for DDI prediction. 
}
\label{flow}
\vspace{-0.3cm}
\end{figure*}

\vspace{-0.2cm}
\subsection{Spectral Clustering}
\label{SC-related}

The fundamental principles of SC posit that data points reside on a manifold within a high-dimensional space. 
SC aims to reveal this underlying manifold through the utilization of Laplacian's eigenvectors \cite{SC02}.
Implementing SC encompasses two pivotal steps: building an undirected graph and cutting the graph \cite{SC01}.
Regarding the first step, all nodes are interconnected.
The edge value between two farther away nodes is lower, while the edge value between two close nodes is higher \cite{SC01}.
The graph-cutting strategy aims to minimize the collective edge values spanning different subgraphs and maximize the cumulative edge values within individual subgraphs. 

Given an undirected graph $\mathcal{G}=({\bf X}, {\bf A})$, where ${\bf X}= \left\{{\bf x}_v\right\}_{v=1}^N \in \mathbb{R}^{N \times d}$ represents the node features, $N$ denotes number of nodes, and $d$ represents the dimension of node features. 
${\bf A} \in \mathbb{R}^{N \times N}$ is the adjacency matrix and can be calculated by many similarity functions.
The matrix element ${A}_{(u,v)}$ represents the pairwise similarity and denotes the edge weight between nodes $u$ and $v$.
The SC method mainly focused on undirected graphs. Thus, the adjacency matrix ${\bf A}$ is symmetric.
The normalized cut (Ncut) \cite{SC02} is the most common way of cutting graphs.
In the Ncut method, the objective function of graph-cutting is denoted as follows:
\begin{equation}
\label{subgraph} 
\mathcal{G}_{\text {Cut }}=\min \frac{1}{2} \sum_{i=1}^C \frac{\mathbf{A}\left(\mathcal{S}_{i}, \overline{\mathcal{S}}_{i}\right)}{vol(\mathcal{S}_{i})},
\end{equation}
where $C$ is the number of clusters, $\mathcal{S}_{i}$ is a subgraph via
cutting, and $\overline{\mathcal{S}}_{i}$ is a complement corresponding to $\mathcal{S}_{i}$.
The term $1/2$ aims to avoid considering each edge twice \cite{SC01}. 
$vol(\mathcal{S}_{i})$ denotes the sum of edge weights within $\mathcal{S}_{i}$.
$\mathbf{A}\left(\mathcal{S}_{i}, \overline{\mathcal{S}}_{i}\right)$ represents the cutting weight between $\mathcal{S}_{i}$ and $\overline{\mathcal{S}}_{i}$:
\begin{equation}
\mathbf{A}\left(\mathcal{S}_{i}, \overline{\mathcal{S}}_{i}\right)=\sum_{u \in {\mathcal{S}_{i}}, v \in {\overline{\mathcal{S}}_{i}}}{A}_{(u,v)}.
\label{subgraph2}
\end{equation}

For ease of computation, following \cite{SC02}, equation (\ref{subgraph}) can be reformulated as:
\begin{equation}
\underbrace{\arg \min }_{\bf F} \operatorname{tr}\left({\bf F}^{\top} {\bf D}^{-1 / 2} {\bf L} {\bf D}^{-1 / 2}  {\bf F}\right) \text { s.t. } {\bf F}^{\top} {\bf F}={\bf I}_{C}
\label{subgraph3}
\end{equation}
where ${\bf L} = {\bf D}-{\bf A}$ represents the graph laplacian matrix of graph $\mathcal{G}$, and ${\bf D} = \operatorname{diag}\left( {\bf A}{\bf 1}_N\right)$ denotes the degree matrix.
${\bf I}_{C}$ is an $C$-dimensional identity matrix.

Let ${\bf P} = {\bf F}^{\top} {\bf D}^{-1 / 2} {\bf L} {\bf D}^{-1 / 2}{\bf F}$, ${\bf P}$ can be divided as follows:
\begin{equation}
\begin{aligned}
{\bf P}&={\bf F}^{\top} {\bf D}^{-1 / 2} {\bf D} {\bf D}^{-1 / 2}{\bf F}-{\bf F}^{\top} {\bf D}^{-1 / 2} {\bf A} {\bf D}^{-1 / 2}{\bf F}\\ 
&= {\bf I}_{C}- {\bf F}^{\top} {\bf D}^{-1 / 2}{\bf A}{\bf D}^{-1 / 2}{\bf F}.
\end{aligned}
\label{subgraph4}
\end{equation}
Therefore, Eq. (\ref{subgraph3}) can be rewritten as:
\begin{equation}
\underbrace{\arg \max }_{\bf F} \operatorname{tr}\left({\bf F}^{\top} {\bf D}^{-1 / 2} {\bf A} {\bf D}^{-1 / 2}  {\bf F}\right) \text { s.t. } {\bf F}^{\top} {\bf F}={\bf I}_{C}
\label{subgraph5}
\end{equation}
Herein, implementing graph-cutting (\ref{subgraph}) is transformed into obtaining appropriate ${\bf F}= \left\{{\bf f}_v\right\}_{v=1}^N \in \mathbb{R}^{N \times C}$, where ${\bf f}_v\in \mathbb{R}^{1 \times C}$ is regarded as a cluster assignment associated with node $v$.
Accordingly, the input data ${\bf X}$ can be partitioned into $C$ clusters. 
The most common calculate way for ${\bf F}$ is performing eigen-decomposition on matrix $ {\bf E} = {\bf D}^{-1 / 2} {\bf A} {\bf D}^{-1 / 2}$.
Then, combining the eigenvectors corresponding to the $C$ maximum eigenvalues to form ${\bf F}$.

\vspace{-0.2cm}
\subsection{Drug Feature Extraction} 
\label{DFE}
The extraction of drug features plays a vital role in model training~\cite{DDIMDL}. 
Recently, many knowledge bases have provided various heterogeneous data sources for researchers.
Zhu et al.~\cite{MADRL} utilized a comprehensive set of biochemical attributes, including molecular substructures, pathways, side effects, targets, enzymes, diseases, genes, and phenotypes.
Typically, each biochemical attribute is associated with a set of descriptors, representing a drug as a binary sequence.
For example, carboxyl, hydroxyl, ester, etc., can be seen as the descriptors of molecular substructures.
The elements in the sequence (1 or 0) indicate the presence or absence of the corresponding descriptor \cite{MDF-SA-DDI}.
Nevertheless, sparse binary sequences are usually high-dimensional, resource-intensive, and may lead to the curse of dimensionality \cite{MDF-SA-DDI}. 
To mitigate this challenge, several works \cite{MDF-SA-DDI, DDIMDL} leveraged similarities between drugs to define drugs' features.
The cosine similarity, the most common way, is calculated as follows:
\begin{equation}
\label{cosine} 
\text{cosine}({\bf c},{\bf d}) = \frac{{\bf c} {\bf d}^{T}}{\|{\bf c} \| \|{\bf d}\|}
\end{equation}
where ${\bf c}$ and ${\bf d}$ are the binary sequences of two drugs under a specific biochemical attribute. 
% MDF-SA-DDI\cite{MDF-SA-DDI} computes three types of drug similarities using three biochemical attributes (targets, enzymes, and molecular substructures) and concatenates these similarities to form the features of drugs. 

\section{Proposed Method}
\label{Method}
% This section provides a comprehensive description of the proposed HMGRL.
% In Section~\ref{framework}, we offer an overview of the HMGRL framework.
% Section~\ref{RaSELP} delves into the relation-aware graph structure embedding for drugs.
% Section~\ref{DPFL} introduces the process of multi-source feature learning for DPs.
% Section~\ref{DPFF} elaborates on the multi-view differentiable spectral clustering.
% Finally, Section~\ref{LF} outlines the loss functions utilized for multi-relational DDI prediction.

\subsection{HMGRL Framework}
\label{framework}
The HMGRL is designed for multi-relational DDI prediction. 
Drugs with known DDIs are typically referred to as “known drugs,” while those without known DDIs are called “new drugs.”
Three specific prediction tasks are defined as follows:

$\bullet$ Task 1: Inferring unobserved interaction events between known drugs.

$\bullet$ Task 2: Inferring interaction events between known and new drugs.

$\bullet$ Task 3: Inferring interaction events between new drugs.

% The three tasks are depicted in Fig.~\ref{3tasks}. 
Consequently, Task 1 features the same drugs in both the training and test sets. For Task 2, the test set shares half of its drugs with the training set. Meanwhile, Task 3 presents entirely different drugs in its training and test sets. Consequently, the prediction challenges escalate progressively from Task 1 through Task 3. Taking Task 2 as an example, the framework of our HMGRL is shown in Fig~\ref{flow}. The primary goal of the proposed HMGRL is to predict the type for each DDI within the test set.
% This task can be considered a multi-classification problem.

\begin{figure*}[t]
\centering 
\includegraphics[width=7.1in,height=2.25in]{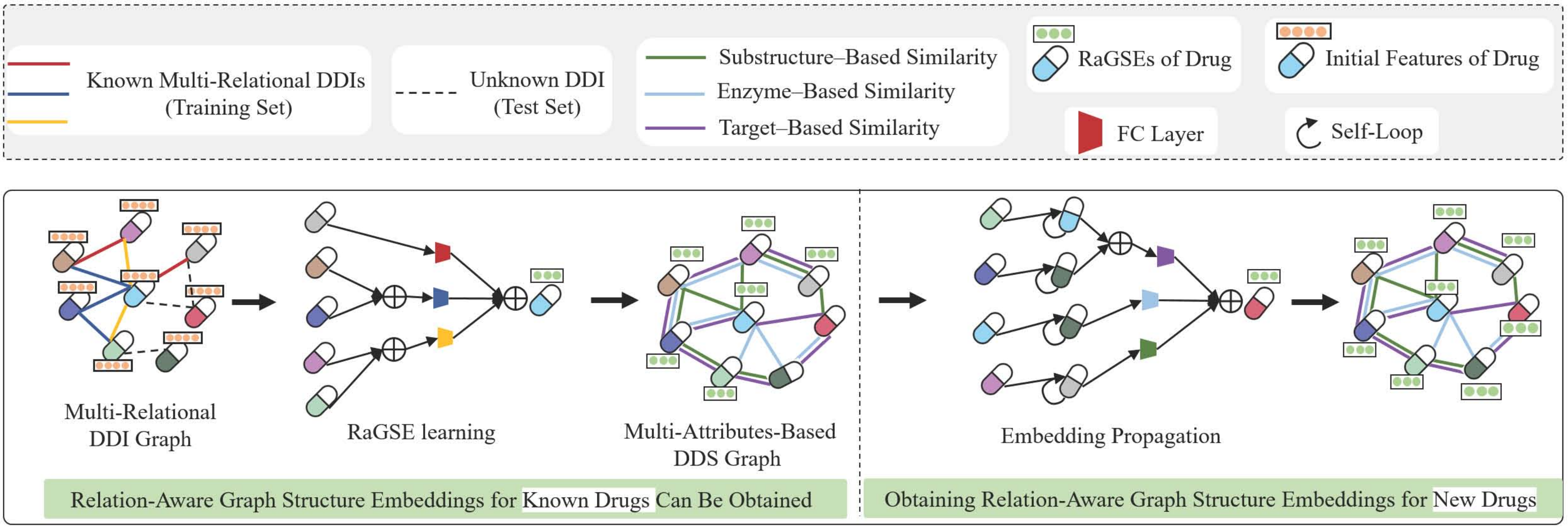}
\vspace{-0.3cm}
\caption{The generating process for drug RaGSEs.
Initially, obtaining RaGSEs for known drugs by aggregating features of neighbors under different relations in the DDI graph.
Following this, RaGSEs for new drugs are obtained by aggregating features of n-hop neighbors under different similarity relations in the DDS graph.
Through this operation, we ensure that all drugs, including new drugs, are equipped with powerful RaGSEs.}
\label{RaGSE}
\vspace{-0.3cm}
\end{figure*}

\subsection{Relation-Aware Graph Structure Embedding }
\label{RaSELP}

Fig.~\ref{RaGSE} exhibits the generating process of RaGSEs for drugs.
Let the multi-relational DDI graph $\mathcal{G}=(\mathcal{V}, \mathcal{E}, \mathcal{R})$, where nodes represent drugs, and edges denote labeled interactions. 
Herein, $\mathcal{V}$ represents the set of drugs, $\mathcal{E}$ denotes the set of labeled interactions, and $\mathcal{R}$ represents the set of interaction event types. 
$R=|\mathcal{R}|$ indicating the number of interaction types.
Let ${\bf X}= \left\{{\bf x}_v\right\}_{v=1}^N \in \mathbb{R}^{N \times d}$ be the initial feature matrix of drugs.
$N$ corresponds to the number of drugs (including known and new drugs), and $d$ denotes the dimensions of initial feature vectors, respectively.  
Previous methods \cite{DDIMDL} and \cite{MDF-SA-DDI} declared that the combination of chemical substructures, targets, and enzymes could achieve the best performance. Therefore, our HMGRL also employs these three biochemical attributes to calculate drug similarities.
Following several experiments, we choose the cosine similarity function, which emerged as the most effective in our evaluations. The calculated similarities are subsequently concatenated to create the initial drug feature matrix, ${\bf X}$.
Let ${\bf A}_r$ $(r = 1, \dots, R)$ correspond to the adjacency matrix under interaction type $r$.
The element of ${\bf A}_{r}$ is denoted by $A_{r(u,v)}$ $(u, v = 1, \dots, N)$, with $A_{r(u,v)} = 1$ if drug $u$ interacts with drug $v$ under type $r$ and $A_{r(u,v)} = 0$ otherwise.

Subsequently, RaGSEs of drug $v$ are generated by Eq.~(\ref{rgcn}).
Going through all the drugs, RaGSEs for all drugs can be obtained: ${\bf \overline X}= \left\{{\bf\overline x}_v\right\}_{v=1}^N \in \mathbb{R}^{N \times d'}$.
However, new drugs, which lack neighbors in the DDI graph, are unable to capture relation-aware graph structural information. 
A widely held assumption posits that drugs sharing similar properties are likely to interact with the same drugs \cite{DDIMDL, MDF-SA-DDI}. In light of this, our HMGRL model allows new drugs to augment their RaGSEs by aggregating the RaGSEs of similar drugs, thereby enhancing their predictive capabilities in the DDI network.

Giving a multi-attribute-based DDS graph $\mathcal{G}_{s}=({\bf \overline X}, \mathcal{A})$, where nodes represent drugs, and edges denote cosine similarities between drugs under different attributes (i.e., targets, enzymes, and molecular substructures).
Let $\mathcal{A}=\left\{{\bf A}_t, {\bf A}_e, {\bf A}_s\right\}\in \mathbb{R}^{N \times N \times 3}$ be the adjacency tensor, where ${\bf A}_t$, ${\bf A}_e$, and ${\bf A}_s$ focus on target, enzyme, and molecular substructure similarities, respectively. 
Let $\hat{\bf A}_{t}$, $\hat{\bf A}_{e}$, and $\hat{\bf A}_{s}$ be the normalized adjacency matrices, and the normalized method is as same as Eq. (\ref{norm}).
To enable new drugs to aggregate the RaGSEs of similar drugs, we design a propagation procedure as follows:
\begin{equation}
{\bf \overline x}_{t,v}^{(l)}=\sum_{u \in \mathcal{N}_v^t} \hat{A}_{t(u,v)} {\bf \overline x}_{t,u}^{(l-1)}  
\label{ddsgcn1}
\end{equation}
\begin{equation} 
{\bf \overline x}_{e,v}^{(l)}=\sum_{u \in \mathcal{N}_v^e} \hat{A}_{e(u,v)} {\bf \overline x}_{e,u}^{(l-1)} 
\label{ddsgcn2}
\end{equation}
\begin{equation} 
{\bf \overline x}_{s,v}^{(l)}=\sum_{u \in \mathcal{N}_v^s} \hat{A}_{s(u,v)} {\bf \overline x}_{s,u}^{(l-1)} 
\label{ddsgcn3}
\end{equation}
where $\hat{A}_{t(u,v)}$, $\hat{A}_{e(u,v)}$, and $\hat{A}_{e(u,v)}$ represent the elements of $\hat{\bf A}_{t}$, $\hat{\bf A}_{t}$, and $\hat{\bf A}_{t}$, respectively. 
$l = 1, \dots, L$, where $L$ signifies the maximum range of embedding propagation. Given that, ${\bf \overline x}_{t,v}^{(0)} = {\bf \overline x}_{e,v}^{(0)}={\bf \overline x}_{s,v}^{(0)} = {\bf \overline x}_v$. Following $L$ rounds of embedding propagation, effective drug RaGSEs are generated in the following way:
\begin{equation}
{\bf \hat x}_v= \sigma\left({\bf \overline x}_{t,v}^{(L)}{\bf W}_t + {\bf \overline x}_{e,v}^{(L)} {\bf W}_e + {\bf \overline x}_{s,v}^{(L)} {\bf W}_s\right)\\
\label{drugembedding}
\end{equation}
where ${\bf W}_t$, ${\bf W}_e$ and ${\bf W}_s \in \mathbb{R}^{d' \times d'}$ are optimizable weight matrices.
This process enables the model understanding of various explicit drug relationships and ensures all drugs can acquire effective RaGSEs.

\vspace{-0.3cm}
\subsection{Multi-Source Feature Learning}
\label{DPFL}

Multiple data sources offer a richer perspective for learning DP features.
We utilize five drug features to boost data diversity and enhance the model's generalization capabilities: RaGSEs, targets, enzymes, molecular substructures, and SMILES strings. 
Fig.~\ref{flow} includes the learning process of comprehensive DP features within multi-source feature learning.

The SMILES string is a line notation that employs a specific set of rules to depict the structure of chemical compounds, with each drug possessing a unique SMILES string \cite{IIFDTI}. 
Within a SMILES string, individual characters symbolize chemical atoms or bonds, and the lengths of these strings are unfixed \cite{IIFDTI}.
This paper transforms each SMILES string into a $p\times q$ feature matrix for easy analysis.
$p$ corresponds to the number of character types appearing in the SMILES strings, and $q$ represents a standardized length for the SMILES string \cite{DeepPurpose}.
Given that the original lengths of the SMILES strings are unfixed, adjustments are made to conform to the unified length of $q$.
Specifically, any portion exceeding the designated length is truncated, and zero-padding is applied if the actual length falls short of $q$. 
Through abundant experiments, the previous method~\cite{DeepPurpose} found that the optimal settings for $p$ and $q$ are $64$ and $100$, respectively. Therefore, we also refer to this configuration.

To mitigate computational overhead, we adopt a batch-wise scheme for training the model and applying it to large-scale data. 
Let $K$ represent the number of DPs in a batch.
Given a DP $k$ (where $k=1,\dots\, K)$, involving drugs $u$ and $v$ (with $u,v=1,\dots\,N, $ and $u \neq v)$.
Let ${\bf{S}}_u$ and ${\bf{S}}_v\in \mathbb{R}^{p \times q}$ be the SMILES-based feature matrices for drugs $u$ and $v$.
We employ a multi-layered 1-D convolutional neural network (CNN) succeeded by a global max-pooling layer to encode the SMILES-based features.
The structure detailed for the 1-D CNN can be found in \cite{DeepPurpose}. 
The DP features derived from SMILES strings are generated as follows:
\begin{equation}
\label{cnnn} 
\widetilde{\bf h}_{smi,k} = \operatorname{CNN}(\mathbf{S}_u||\mathbf{S}_v,\Theta_{\mathrm{CNN}})\\
\end{equation}
where $\widetilde{\bf h}_{smi,k}\in \mathbb{R}^{1 \times d^{att}}$, and $d^{att}$ denote the output dimension of the CNN. $||(\cdot)$ represents the concatenate operation. 
$\Theta_{\mathrm{CNN}}$ represent trainable weight parameters.

In addition, we create four encoders for coding the remaining four features. Each encoder incorporates a fully connected (FC) layer and a transformer layer \cite{transformer1}.
Vectors ${\bf {\hat x}}_{u}$ and ${\bf {\hat x}}_{v}$ represent the RaGSEs for drugs $u$ and $v$, respectively. 
Consequently, the encoding of DP features derived from the RaGSE source is structured as follows:
\begin{equation}
\label{dp}
\widetilde{\bf h}_{emb,k} = \operatorname{Encoder 1}({\bf \hat x}_{u}||{\bf \hat x}_{v},\Theta_{\mathrm{En1}})\\ 
\end{equation}
where $\widetilde{\bf h}_{emb,k}\in \mathbb{R}^{1 \times d^{emb}}$, $d^{emb}$ is the output dimension of Encoder 1, and $\Theta_{\mathrm{En1}}$ represent trainable weight parameters.

Vectors ${\bf a}_{t,u}$ and ${\bf a}_{t,v}$ represent the target similarities of drugs $u$ and $v$, corresponding to the vectors in $u$-th and $v$-th row of ${\bf A}_t$, respectively.
Similarly, ${\bf a}_{e,u}$ and ${\bf a}_{e,v}$ are enzyme similarities of these drugs and derive from ${\bf A}_e$. Lastly, ${\bf a}_{s,u}$ and ${\bf a}_{s,v}$ denote the substructure similarities and come from ${\bf A}_s$.
Accordingly, these three features are encoded as follows:
\begin{equation}
\label{dpembed} 
\widetilde{\bf h}_{tar,k} = \operatorname{Encoder2}({\bf a}_{t,u}||{\bf a}_{t,v},\Theta_{\mathrm{En2}})\\ 
\end{equation}
\begin{equation}
\label{dptar} 
\widetilde{\bf h}_{enz,k} = \operatorname{Encoder3}({\bf a}_{e,u}||{\bf a}_{e,v},\Theta_{\mathrm{En3}})\\ 
\end{equation}
\begin{equation}
\label{dpenzy} 
\widetilde{\bf h}_{sub,k} = \operatorname{Encoder4}({\bf a}_{s,u}||{\bf a}_{s,v},\Theta_{\mathrm{En4}})\\ 
\end{equation}
where $\widetilde{\bf h}_{tar,k}$, $\widetilde{\bf h}_{enz,k}$ and $\widetilde{\bf h}_{sub,k}\in \mathbb{R}^{1 \times d^{att}}$ represent DP features, and $d^{att}$ is the output dimension of Encoders $2$ to $4$.
$\Theta_{\mathrm{En2}}$, $\Theta_{\mathrm{En3}}$, and $\Theta_{\mathrm{En4}}$ are optimizable weight parameters.

All coded results are concatenated to serve as the comprehensive DP features:
\begin{equation}
\label{dpcom}  
\widetilde{\bf h}_{k} = || \left(\widetilde{\bf h}_{smi,k},\widetilde{\bf h}_{emb,k},\widetilde{\bf h}_{tar,k},\widetilde{\bf h}_{enz,k},\widetilde{\bf h}_{sub,k}\right)
\end{equation}

% \vspace{-0.2cm}
\subsection{Multi-View Differentiable Spectral Clustering}
\label{DPFF}

The MVDSC module is developed to capture implicit correlations between DPs from multiple views. 
The MVDSC module consists of four DSCs, and each DSC pays attention to distinct implicit correlations.
DSC 1 focuses on comprehensive features, and DSC 2 to 4 emphasizes targets, enzymes, and substructures, respectively. 
Four DSCs have different inputs but have identical construction. 
Fig. \ref{flow} includes the learning process of high-level DP representations through MVDSC.

\begin{figure}[t]
\centering
\includegraphics[width=3.5in,height=1.95in]{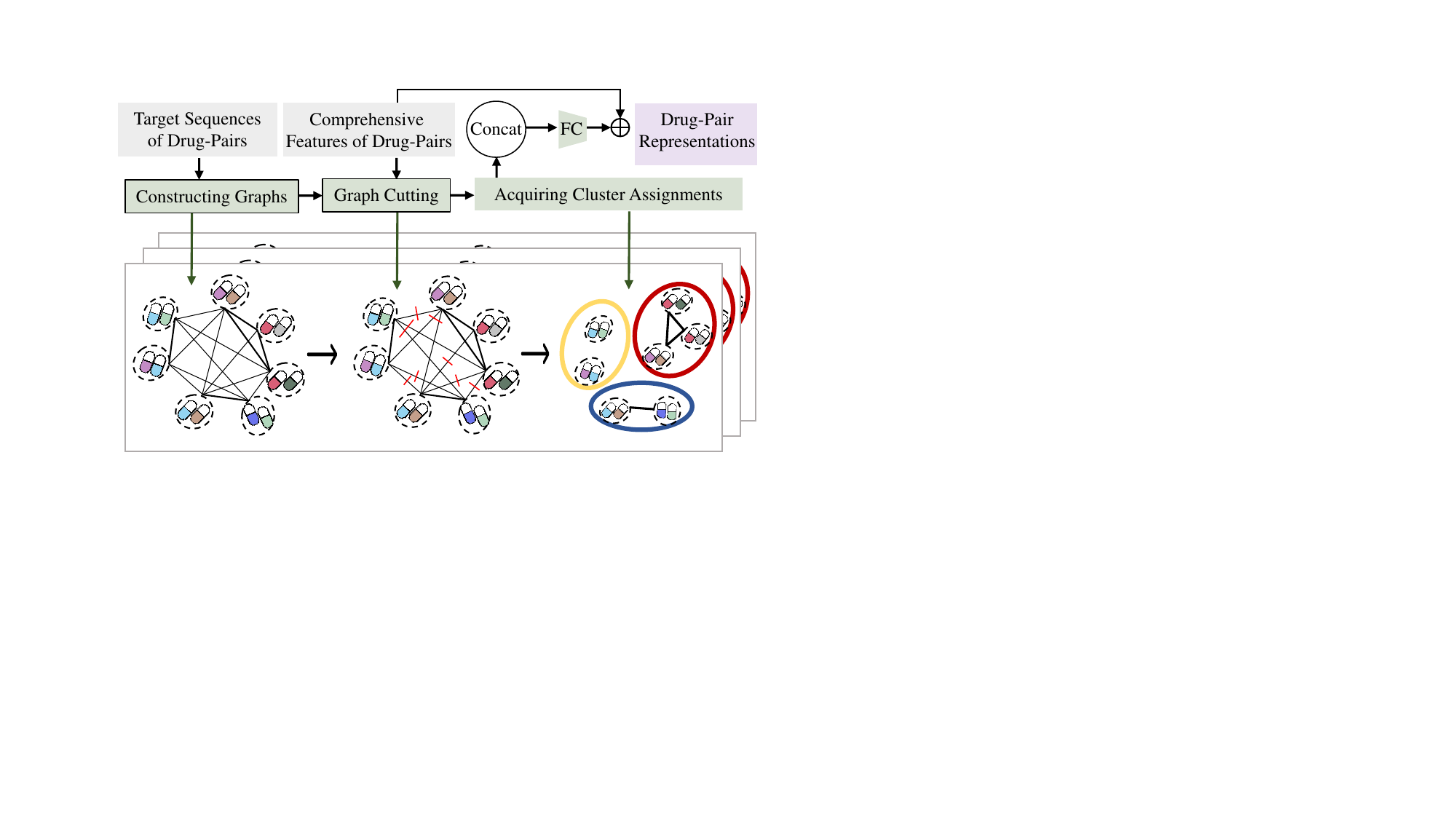}
\vspace{-0.4cm}
\caption{The obtaining process for DP representations within DSC 2.}
\label{MHDSC3}
\vspace{-0.5cm}
\end{figure}

Using DSC $2$ as an example, this module takes target sequences and comprehensive DP features to produce DP representations focusing on target-based correlations.
Fig. \ref{MHDSC3} shows the obtaining process for DP representation within DSC $2$.
Target sequences of DPs are utilized to establish multiple undirected graphs, where nodes denote DPs, and edges exist only between DPs with one or more of the same target descriptors. Therefore, these graphs are not fully connected.
Let ${\bf T}= \left\{{\bf t}_k\right\}_{k=1}^K \in \mathbb{R}^{K \times T} $ represent the target sequences of DPs within a batch.
${\bf t}_k\in \mathbb{R}^{1 \times T}$ is an element-wise summation result of the binary target sequences of two drugs associated with DP $k$, and 
$T$ stands for the count of target descriptors.
$\widetilde{\bf H}= \left\{\widetilde{\bf h}_{k}\right\}_{k=1}^K $ denote the DP comprehensive features that derived from Eq. (\ref{dpcom}).

Based on the principles outlined in SC \cite{SC0}, we have designed a two-step process for each DSC: graph construction and graph cutting.
In DSC $2$, target sequences ${\bf T}$ is utilized to construct graphs as follows:
\begin{equation} 
\hat {\bf T}_{(m)}= {\bf T}{\bf W}_{T(m)} \\ 
\label{dsc1}
\end{equation}
\begin{equation}  
{\bf A}_{(m)}=\operatorname{softmax}\left( \hat {\bf T}_{(m)}\hat {\bf T}_{(m)}^{\top} \right)\in {\mathbb R}^{K \times K}  
\label{dsc2}
\end{equation}
where ${\bf A}_{(m)}$ can be seen as a normalized adjacency matrix, and each element in ${\bf A}_{(m)}$ denotes the edge weight between two corresponding DPs.
${\bf W}_{T(m)}$ denote a trainable weight matrix.
In this context, $m= 1,2,\dots, M$, yielding a total of $M$ adjacency matrices.
This setting draws inspiration from the multi-head strategy, which can stabilize the learning process of self-attention \cite{GAT+chapter2018}.

Proceeding further, we conduct graph cutting leveraging $\widetilde{\bf H}$ in the subsequent manner:
\begin{equation}
{\bf F}_{(m)}=\sigma\left({\bf A}_{(m)}\widetilde {\bf H}{\bf W}_{H(m)} \right)\in \mathbb{R}^{K \times C}
\label{dsc4}
\end{equation}
where ${\bf W}_{H(m)}$ denote a trainable weight matrix.
The matrix ${\bf F}_{(m)}$ represents cluster assignments for DPs, while $C$ signifies the number of clusters. The activation function $\sigma(\cdot)$ is the ReLU function, guaranteeing that the cluster assignments remain non-negative.
The multi-head strategy makes for multiple cluster assignments.
All these cluster assignments are concatenated to create DP representations:
\begin{equation}
\label{dptarget}  
{\bf F}=||\left({\bf F}_{(1)},{\bf F}_{(2)},\dots,{\bf F}_{(M)}\right)
\end{equation}
\begin{equation}
{\bf F}_{T}= \sigma \left({\bf F}{\bf W}_{T}+\widetilde{\bf H}\right)
\label{dsc5}
\end{equation}
where ${\bf F}_{T}$ serves as the output of DSC 2 and denotes DP representations that focus on target-based correlations.
${\bf W}_{T}$ denotes the trainable weight matrices.

% According to Eq. (\ref{subgraph5}), the objective function is intuitively defined to minimize:
% \begin{equation}
% \begin{aligned}
% \mathcal{L} = &-\operatorname{Tr}\left({\bf F}_{(m)}^{\top} { \bf D}^{-1/2} {\bf A}_{(m)} { \bf D}^{-1/2}{\bf F}_{(m)}\right)\\
% &+ \left({\bf F}_{(m)}^{\top}{\bf F}_{(m)}- \mathbf{I}_C\right)
% \end{aligned}
% \label{lossgc0}
% \end{equation}

To direct and constrain the model's training toward achieving optimal cluster assignments, we introduce two unsupervised loss functions.
The two loss functions are derived from the objective function of the Ncut \cite{SC2,SCGNN}, Eq. (\ref{subgraph5}).
The first loss function is designed to encourage minimizing the weights between clusters and maximizing the weights within clusters:
\begin{equation}
\begin{aligned}
\mathcal{L}_{gc2}&= -\frac{1}{M}\sum_{m=1}^{M}\frac{\operatorname{Tr}\left({\bf F}_{(m)}^{\top} { \bf D}^{-1/2} {\bf A}_{(m)} { \bf D}^{-1/2}{\bf F}_{(m)}\right)}{\operatorname{Tr}\left({\bf F}_{(m)}^{\top}{ \bf D}^{-1/2} { \bf D}{ \bf D}^{-1/2} {\bf F}_{(m)}\right)}\\
&= -\frac{1}{M}\sum_{m=1}^{M}\frac{\operatorname{Tr}\left({\bf F}_{(m)}^{\top} {\bf A}_{(m)} {\bf F}_{(m)}\right)}{\operatorname{Tr}\left({\bf F}_{(m)}^{\top} {\bf F}_{(m)}\right)}
\label{lossgc}
\end{aligned}
\end{equation}
where $\mathbf{D}$ represents the degree matrix of $\bf{A}_{(m)}$.
Given that ${\bf A}_{(m)}$ has been normalized through the softmax function as depicted in Eq. (\ref{dsc2}), $\bf{D}$ becomes the $K$-dimensional identity matrix.
According to \cite{SCGNN}, $\mathcal{L}_{gc2}$ attains its maximum value of 0 when $\operatorname{Tr}\left({\bf F}_{(m)}^{\top} {\bf A}_{(m)} {\bf F}_{(m)}\right) =0 $.  This scenario arises when, for each pair of interconnected nodes (i.e., ${A}_{(m)(k,q)}>0$), the cluster assignments are orthogonal, meaning $\left\langle{\mathbf f}_{k}, {\mathbf f}_{q}\right\rangle=0$.
Conversely, $\mathcal{L}_{gc2}$ drops to its minimum value, -1, when $\operatorname{Tr}\left({\bf F}_{(m)}^{\top} {\bf A}_{(m)} {\bf F}_{(m)}\right)= \operatorname{Tr}\left({\bf F}_{(m)}^{\top} {\bf F}_{(m)}\right)$.
This situation emerges when, in a graph with $C$ disconnected clusters, cluster assignments are consistent for all nodes within the same cluster and remain orthogonal to the assignments of nodes across varying clusters.

Simultaneously, the second essential loss is integrated to ensure that the cluster assignments maintain orthogonality and that clusters remain roughly equivalent in size \cite{SCGNN}. The formulation is in the following way:
\begin{equation}
\mathcal{L}_{or2}=\frac{1}{M}\sum_{m=1}^{M}\left\|\frac{{\bf F}_{(m)}^{\top} {\bf F}_{(m)}}{\left\|{\bf F}_{(m)}^{\top} {\bf F}_{(m)}\right\|_F}-\frac{\mathbf{I}_C}{\sqrt{C}}\right\|_F
\label{lossor} 
\end{equation}
where $\mathbf{I}_{C}$ denotes a $C$-dimensional identity matrix, and $\left\| \cdot
\right\|_F$ indicates the Frobenius norm.

Finally, we combine the two unsupervised loss functions to form the regularization term within DSC 2:
\begin{equation}
\mathcal{L}_{DSC2} = \mathcal{L}_{gc2} +\mathcal{L}_{or2}
\label{dsc22} 
\end{equation}
where $\mathcal{L}_{DSC2}$ enhances the model's performance, ensuring its alignment with the core clustering objectives.

DSC 1 emphasizes comprehensive features. 
Therefore, DSC 1 utilizes $\widetilde{\bf H}$ to construct graphs, not ${\bf T}$.
Accordingly, the normalized adjacency matrix is defined as ${\bf S}_{(m)} = \operatorname{softmax}\left(\hat {\bf H}_{(m)}\hat {\bf H}_{(m)}^{\top} \right)$, where $\hat {\bf H}_{(m)}= \widetilde{\bf H}{\bf W}_{H(m)}$, and ${\bf W}_{H(m)}$ is a trainable weight matrix.
With $m$ ranging from $1$ to $M$, the multi-head strategy remains integral to DSC 1. 
Subsequent steps mirror those of DSC 2.
The output of DSC 1 is denoted as ${\bf F}_{H}$, denoting DP representations focus on comprehensive features-based correlations.
Additionally, within the DSC 1, the regularization terms given by $\mathcal{L}_{DSC1} = \mathcal{L}_{gc1} +\mathcal{L}_{or1}$ and also need to be accounted for.

DSC 3 and 4 pay attention to enzyme and substructure features, respectively.
Consequently, they use the corresponding enzyme and substructure sequences for graph construction.
Herein, the enzyme or substructure sequences for a given DP are derived from the element-wise summation of the binary enzyme or substructure sequences of the two drugs implicated in that DP.
The outputs of DSC 3 and DSC 4 are represented as ${\bf F}_{E}$ and ${\bf F}_{S}$, spotlighting DP representations centered on enzyme-related and substructural correlations, respectively.
In alignment with this, it's imperative to factor in the regularization terms for both DSC 3 and DSC 4, denoted as $\mathcal{L}_{DSC3}$ and $\mathcal{L}_{DSC4}$.
 
Subsequently, these outputs of DSC 1 to 4 are concatenated to generate the high-level representations for DPs:
\begin{equation}
\hat {\bf F} = \|\left({\bf F}_{H},{\bf F}_{T},{\bf F}_{E},{\bf F}_{S}\right) .
\label{high-level}
\end{equation}
Meanwhile, their regularization terms are also comprehensively considered as follows:
\begin{equation} 
\mathcal{L}_{DSC} = \frac{1}{4}\left(\mathcal{L}_{DSC1}+\mathcal{L}_{DSC2}+\mathcal{L}_{DSC3}+\mathcal{L}_{DSC4}\right)   
\label{dscloss}
\end{equation}

\begin{algorithm}[]  % % \\tiny\\scriptsize\\footnotesize\\small\\normalsize
	\caption{Pseudocode of the proposed HMGRL.}\label{pseudocode}%算法名字
    \label{Pseudocode} 
	\LinesNumbered %要求显示行号
	\KwIn{$N_{epochs}$:~maximum epochs; $N_{batches}$:~total batch; $\mathcal{T}_{r}$: the set of training DDIs; $\bf{\hat{Y}}$:~ground truth of training DDIs; $\mathcal{T}_{e}$: the set of test DDIs.}%输入参数
	\KwOut{${\bf Y}_{te}$: predicted types of test DDIs.}%输出
    $\bf{X} \leftarrow$ Obtain drugs features via (\ref{cosine}). \\
    $\mathcal{G} \leftarrow$ Construct a DDI graph using known DDIs and $\bf{X}$. \\ 
    {//$\ast$ \textit{Training Procedure} $\ast$//} \\
    \For{ $ epoch \in [1,N_{epochs}]$}{
 	\For{ $ i \in [1,N_{batch}]$}{
    ${\bf \overline X} \leftarrow $ Use $\mathcal{G}$ to learn RaGSEs for known drugs via (\ref{rgcn}). \\ 
    $\mathcal{G}_{s} \leftarrow$ Construct a DDS graph using cosine similarities of drugs and ${\bf \overline X}$. \\ 
    ${\bf \hat X} \leftarrow $ Use $\mathcal{G}_{s}$ to learn RaGSEs for all drugs via (\ref{ddsgcn1}) to (\ref{drugembedding}). \\  
    Obtain $i$-th batch training DDIs from $\mathcal{T}_{r}$ \\ 
    $\widetilde{\bf{H}} \leftarrow$ Obtain comprehensive features of DPs via (\ref{cnnn}) to (\ref{dpcom}).\\ 
    ${\bf F}_{T} \leftarrow$ Obtain DP representations using DSC 2 via (\ref{dsc1}) to (\ref{dsc5}).\\
    $\mathcal{L}_{DSC2} \leftarrow $ Derive regularization term via (\ref{lossgc})to (\ref{dsc22}).\\
    ${\bf F}_{H}, {\bf F}_{E},$ and ${\bf F}_{S}\leftarrow $ Obtain DP representations via DSC 1, 3, and 4. \\
    $\hat{{\bf F}} \leftarrow$ Obtain high-level DP representations via (\ref{high-level}).\\
    $\mathcal{L}_{DSC} \leftarrow$ Obtain the total regularization term via (\ref{dscloss}).\\ 
    $\mathbf{Y} \leftarrow $ Obtain probability distributions of DPs using $\hat{{\bf F}}$ via (\ref{predict}).\\
    $\mathcal{L}_{ce} \leftarrow$ Use $\bf{\hat{Y}}$ and $\mathbf{Y}$ generate classification loss via (\ref{lossce}).\\
    $\mathcal{L} \leftarrow$ Obtain the final loss via (\ref{concatloss}).\\
    Minimize $\mathcal{L}$ and update ${\bf \overline X}$, $\hat{\bf X}$, $\widetilde{\bf H}$, ${\bf F}_{H}$, ${\bf F}_{T}$, ${\bf F}_{E}$, and ${\bf F}_{S}$.\\
    }}
    {//$\ast$ \textit{Testing Procedure} $\ast$//} \\
    Predict types of the DDIs in $\mathcal{T}_{e}$ using the trained HMGRL.\\
\end{algorithm}
\vspace{-0.4cm}

\subsection{Loss Function}
\label{LF}
We employ a multi-layer perception (MLP) \cite{MLP} as a decoder that maps the high-level DP representations $\hat {\bf F}$ into the probability distribution space ${\bf Y}$:
\begin{equation}
{\bf Y}=\operatorname{Decoder}\left(\hat {\bf F},\Theta_{\mathrm{MLP}}\right)\in \mathbb{R}^{K \times R}
\label{predict}
\end{equation}
where $\Theta_{\mathrm{MLP}}$ represents the trainable parameters of Decoder, and $R$ denotes the number of DDI event types. 
In the Decoder, the first FC layer is followed by an activation function, ReLu \cite{ReLu}, and a dropout layer \cite{dropout}.
A softmax function succeeds the second FC layer. 

We choose the cross-entropy (CE) loss function as the classification loss function:
\begin{equation}
\mathcal{L}_{ce}= -\sum_{k=1}^{K} \sum_{r=1}^{R} \hat{y}_{k,r} \log \left(y_{k,r}\right)
\label{lossce}
\end{equation}
where $\hat{y}_{k,r}$ represents the $r$-th item of $\hat{\bf{y}}_{k}$, while $\hat{\bf{y}}_{k}$ denotes the ground-truth vector (one-hot encoding) for DP $k$. $K$ signifies the number of training DDI samples in a batch.
$y_{k,r}$, the $r$-th item of ${\bf y}_k$, indicating the predicted probability score of for DP $k$ within class $r$. 
The regularization item, $\mathcal{L}_{DSC}$, is considered an auxiliary loss for the classification loss.
Consequently, the model's total loss function is as follows:
\begin{equation}
\mathcal{L}=\mathcal{L}_{ce}+\alpha \mathcal{L}_{DSC} 
\label{concatloss}
\end{equation}
where $\alpha$ is hyperparameters. 

Our HMGRL presents a semi-supervised approach for DDI prediction. 
Notably, it integrates both known and new drugs throughout each task's training and testing stages. 
For a more nuanced understanding of our objective's optimization process, we provide pseudocode in Algorithm \ref{Pseudocode}.
% Our HMGRL commences with using cosine similarities of drugs under three attributes as drug initial features (Line 1).
% Following this, we construct a multi-relational DDI graph using the training dataset (known DDIs) coupled with the drug's initial features (Line 2).
% Lines 3-21 carry out the training procedure of HMGRL.
% Specifically, Lines 6-8 align with the descriptions in subsections~\ref{RaSELP}, 
% Lines 9-10 correspond to~subsections~\ref{DPFL}, 
% Lines 11-15 reflect to subsections~\ref{DPFF}, 
% and Lines 16-19 correspond to the content in subsections~\ref{LF}. 
% Post-training, the refined HMGRL is used to predict the labels of test DDIs.

\vspace{-0.2cm}
\section{Experimental setup}
\label{experiment}

\subsection{Datasets}\label{dataset}

In this paper, we utilize two large real heterogeneous datasets derived from DrugBank, each comprising three distinct tasks, to evaluate the effectiveness and competitiveness of our HMGRL. 
DrugBank serves as a comprehensive knowledge base that can provide heterogeneous, detailed, up-to-date, quantitative analysis of drugs. Leveraging such extensive and diverse datasets from DrugBank, we can enhance the robustness and applicability of our methodology.

Dataset 1 comprises 572 drugs with 37,264 pairwise DDIs spanning 65 interaction types and collected by Deng et al. \cite{DDIMDL}. 
Each drug is characterized by five attributes: enzymes, targets, pathways, molecular substructures, and SMILES string. The acquisition and processing of the dataset have been thoroughly detailed in the previous study~\cite{DDIMDL}.
 
Dataset 2 is constructed by us. 
It consists of 1000 drugs, each characterized by four distinct features: enzymes, targets, molecular substructures, and SMILES string. 
We obtained 206,029 pairwise DDIs, encompassing 99 event types.
Dataset 2 encompasses more comprehensive DDI event information compared to Dataset 1.  
This expanded data set in Dataset 2 can aid in more effective model training, although the increased variety of DDI event types also introduces challenges in data fitting.
Dataset 2 provides an evaluation platform for assessing performance on a large-scale dataset. 

\begin{figure*}[t]
\centering 
\subfigure[Task 1]{\includegraphics[width=2.2in,height=1.3in]{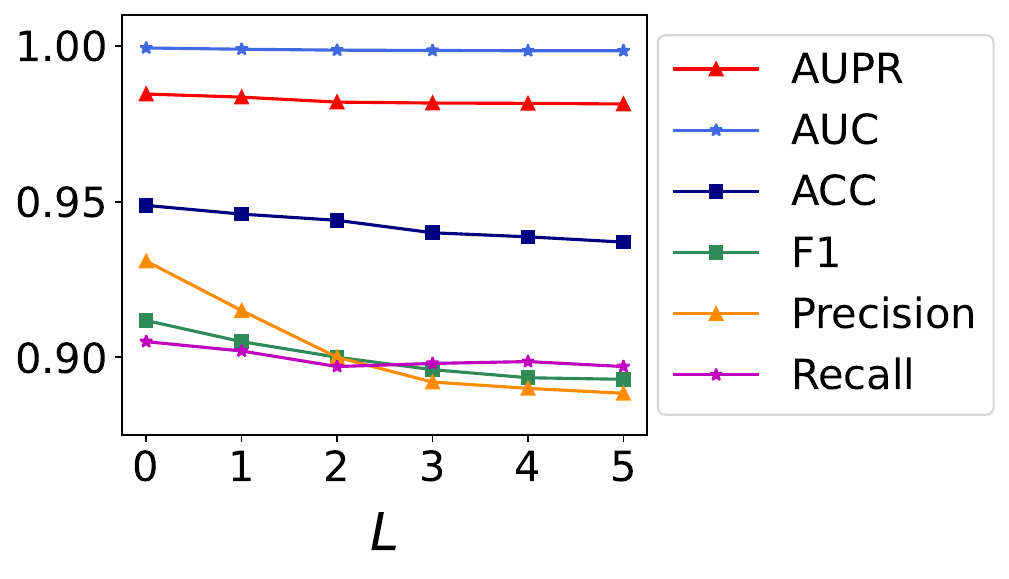}} 
\hspace{2mm}
\subfigure[Task 2]{\includegraphics[width=2.2in,height=1.3in]{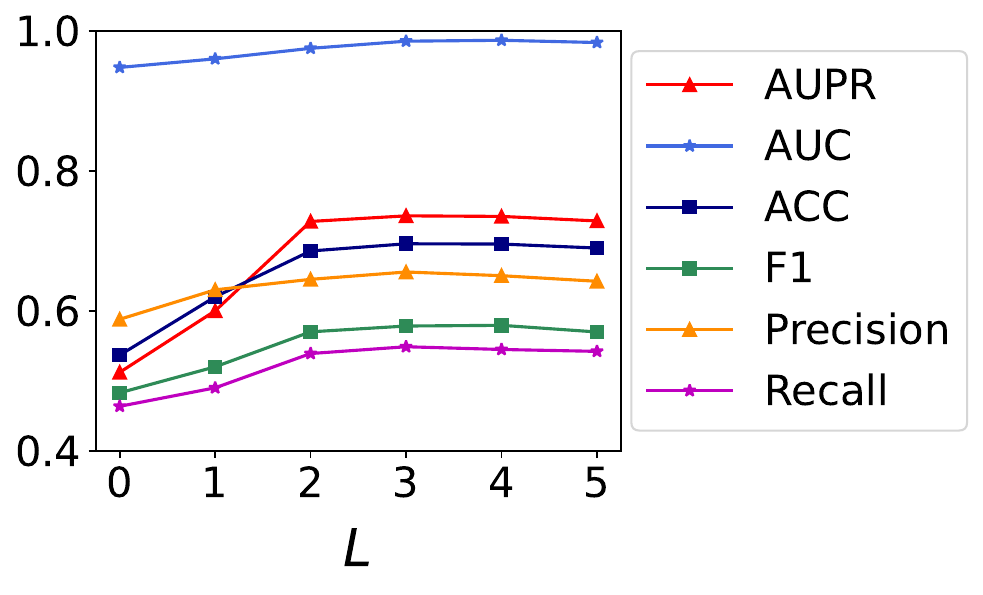}}
\hspace{2mm}
\subfigure[Task 3]{\includegraphics[width=2.2in,height=1.3in]{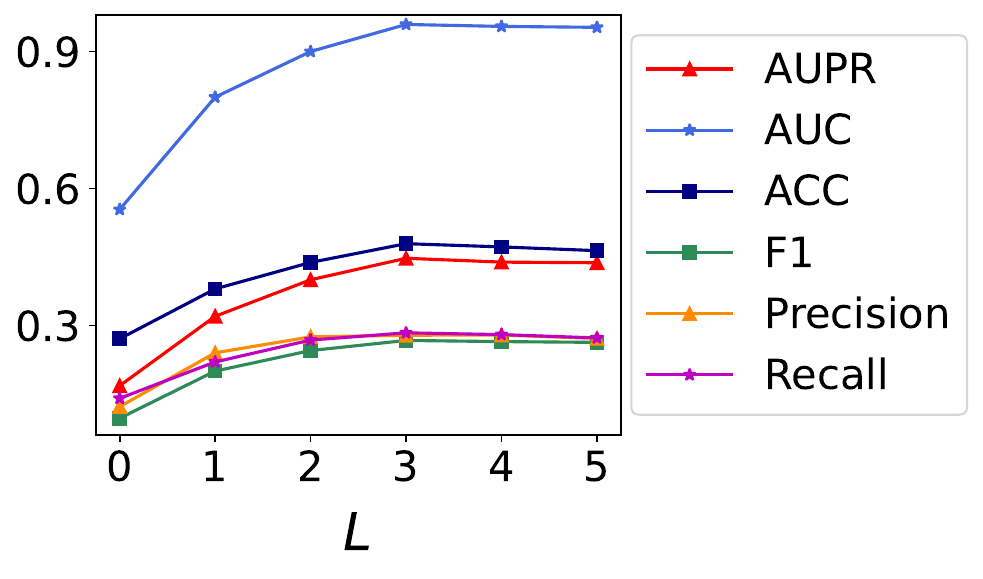} }
\caption{Six metrics versus the embedding propagation distance $L$ on different tasks of Dataset 1.}
\label{n-hop}
\end{figure*}

\begin{figure*}[t]
\centering 
\subfigure[Task 1]{\includegraphics[width=2.1in,height=1.58in]{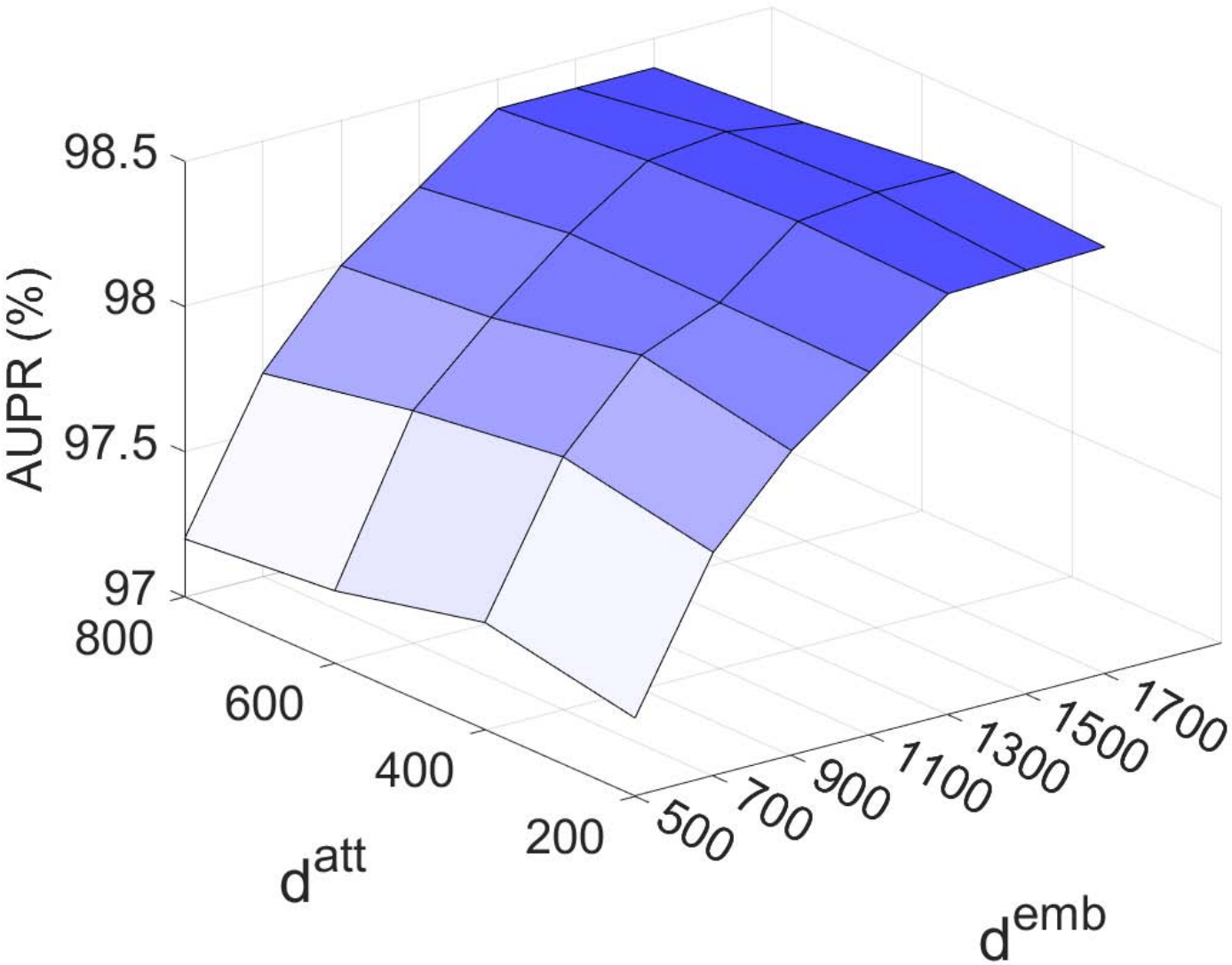}} 
\hspace{1mm}
\subfigure[Task 2]{\includegraphics[width=2.25in,height=1.58in]{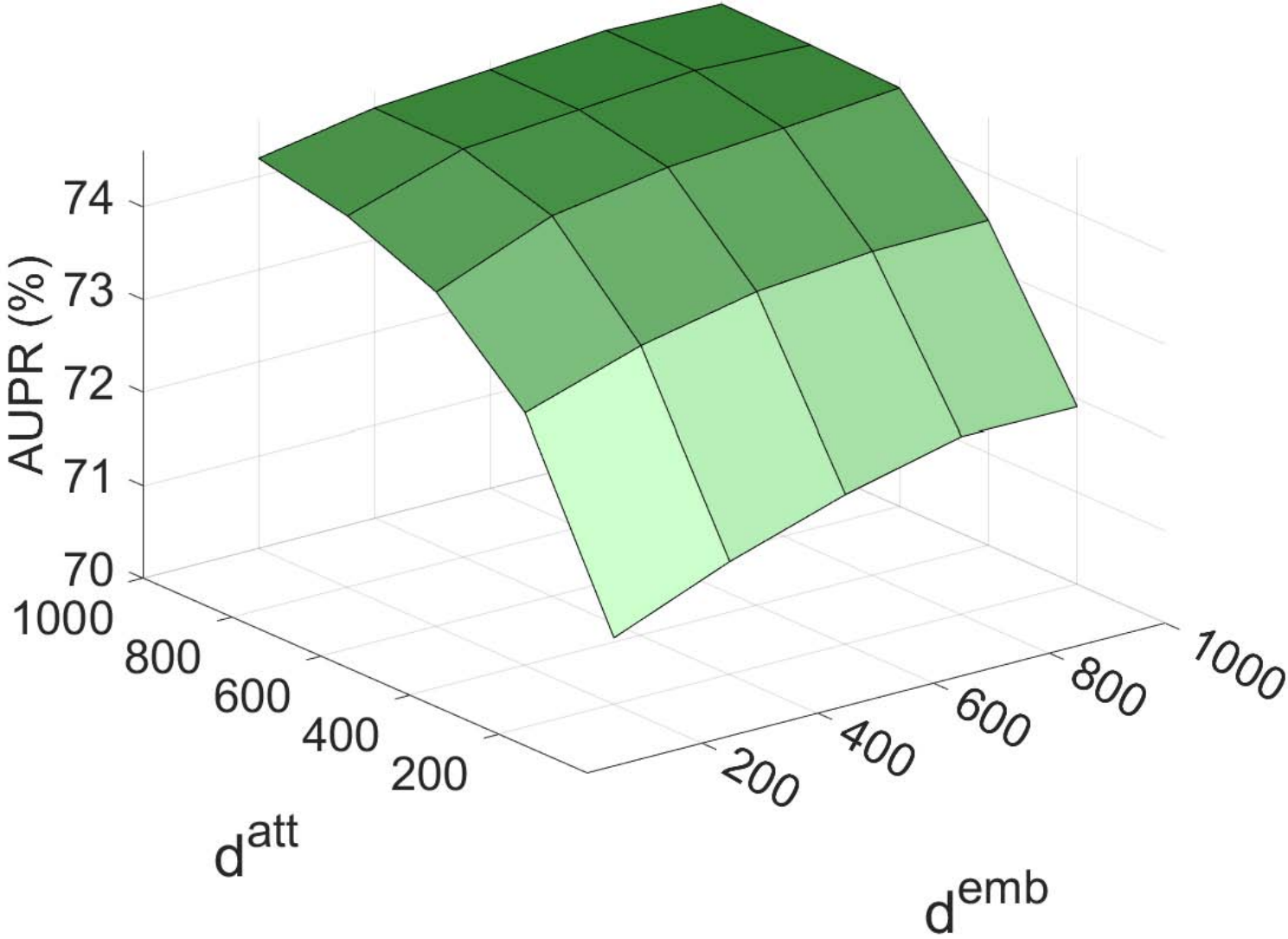}} 
\hspace{1mm}
\subfigure[Task 3]{\includegraphics[width=2.25in,height=1.58in]{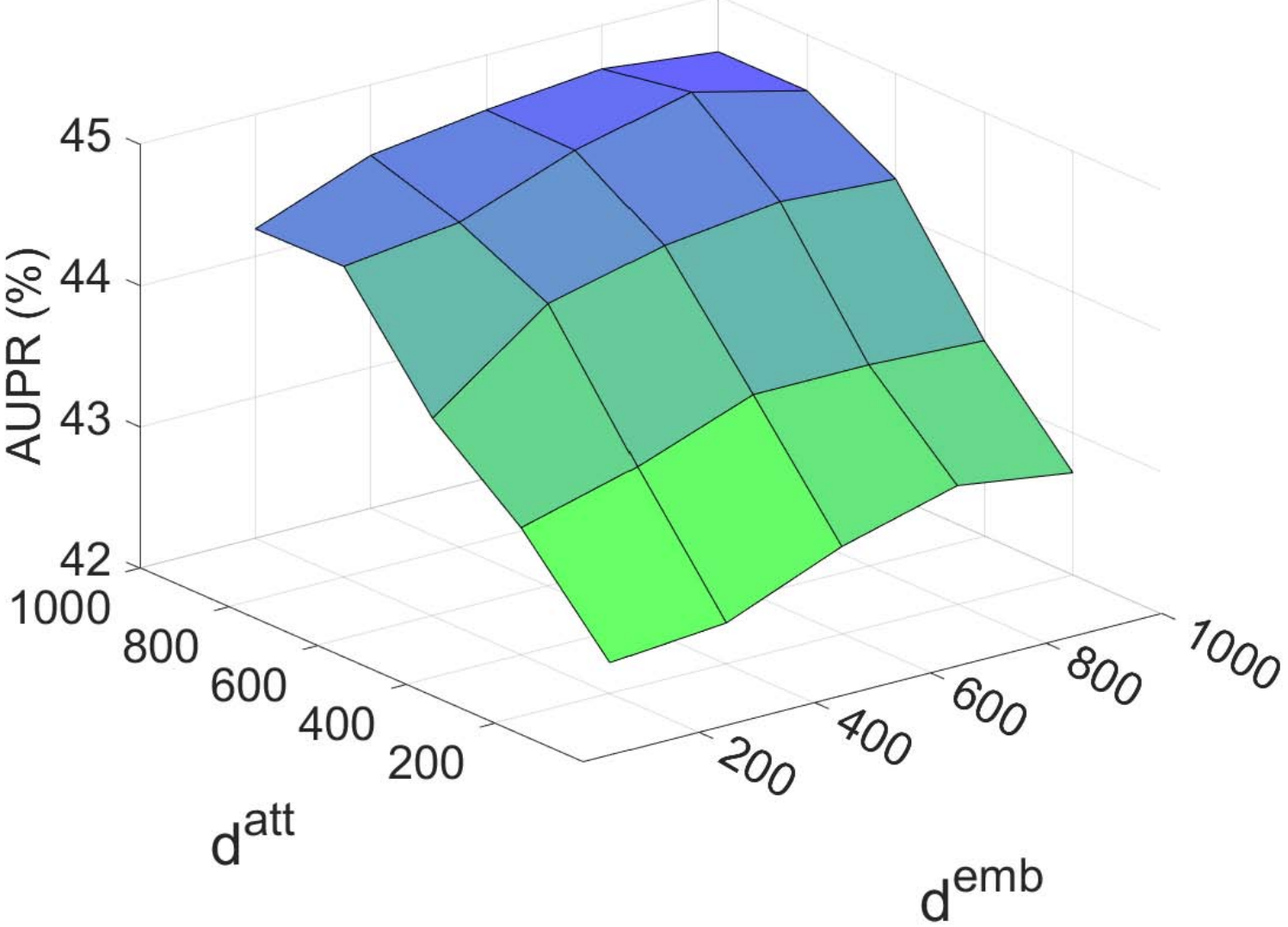}}
\caption{AUPR (\%) versus the dimensions $d^{att}$ and $d^{emb}$ on different tasks of Dataset 1.}
\label{dimtask}
\end{figure*}

\vspace{-0.2cm}
\subsection{Baselines}\label{Baselines}
We compare the proposed HMGRL with five baselines:

$\bullet$ DDIMDL~\cite{DDIMDL} is a DNN model that employs multiply drug-drug-similarities for DDI prediction. 

$\bullet$ RANEDDI~\cite{RANEDDI} is a GNN-based model that constructs a multi-relational DDI graph to learn drug embeddings. 

$\bullet$ MDF-SA-DDI~\cite{MDF-SA-DDI} consists of four encoders to encode DP features and adopts transformers to perform feature fusion to acquire DP representation.

$\bullet$ MAEDDIE~\cite{MAEDDIE} is an attention mechanism-based encoder-decoder model that constructs a multisource feature fusion network to learn drug embeddings for DDI prediction.

$\bullet$ MCFF-MTDDI~\cite{MCFF-MTDDI} is a gated recurrent unit-based multi-channel feature fusion module that utilizes multiple drug-related features to yield comprehensive representations of DPs.

$\bullet$ MM-GANN-DDI~\cite{MM-GANN-DDI} is a novel GNN-based model, effectively leveraging topological information from the DDI graph and efficiently predicting DDI types for new drugs.

\vspace{-0.2cm}
\subsection{Experimental Setting}\label{Experimental}
We evaluate our HMGRL's performance by conducting Tasks 1, 2, and 3. 
In Task 1, we employ 5-fold cross-validation (5-CV) to divide DDIs into five subsets, using four subsets for training and the remaining one for testing.
For Tasks 2 and 3, we partition the drugs, rather than the DDIs, into five subsets using 5-CV. Four subsets are taken as known drugs, while the remaining subset is designated as new drugs.
In Task 2, we take the DDIs involving two known drugs as the training set while taking DDIs containing one known drug and one new drug as the test set. 
In Task 3, the training set remains identical to Task 2, while the test set only involves new drugs.

We assess our model using six metrics: 
the area under the precision-recall curve (AUPR), the area under the ROC curve (AUC), accuracy (ACC), F1 score, Precision, and Recall. 
The activation function and dropout layer are employed between the FC layers. 
The ReLU activation function and the Radam optimizer \cite{Radam} are leveraged by default.
Furthermore, we integrate a data augmentation algorithm, Mixup~\cite{mixup}, into our HMGRL.
Mixup amplifies the original data volume and enhances the model's generalization capability and robustness by intermixing data points. 
A comprehensive exploration of Mixup's mechanics and advantages can be found in previous work \cite{mixup}.
The proposed HMGRL is implemented with the deep learning library PyTorch.
The Python and PyTorch versions are 3.8.10 and 1.9.0, respectively.
All experiments are conducted on a Windows server with a GPU (NVIDIA GeForce RTX 4090).

\begin{figure*}[t]
\centering 
\subfigure[Task 1]{\includegraphics[width=2.15in,height=1.4in]{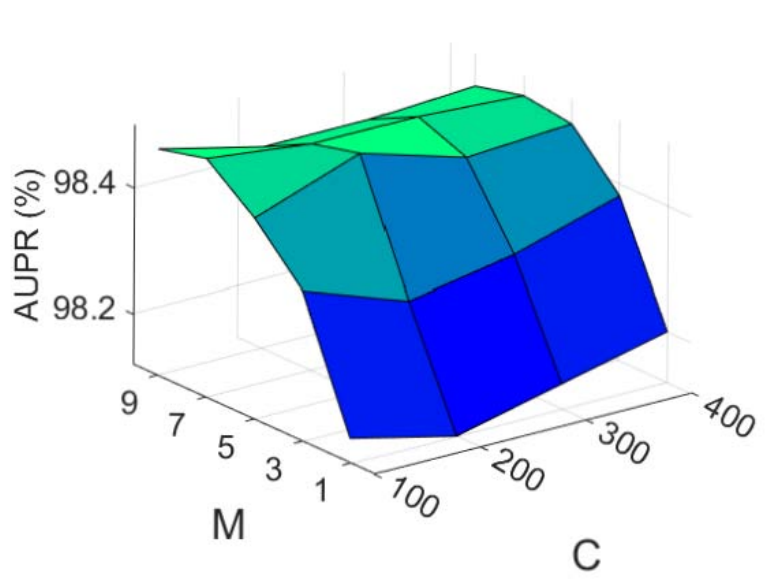}} 
\hspace{3mm}
\subfigure[Task 2]{\includegraphics[width=2.15in,height=1.4in]{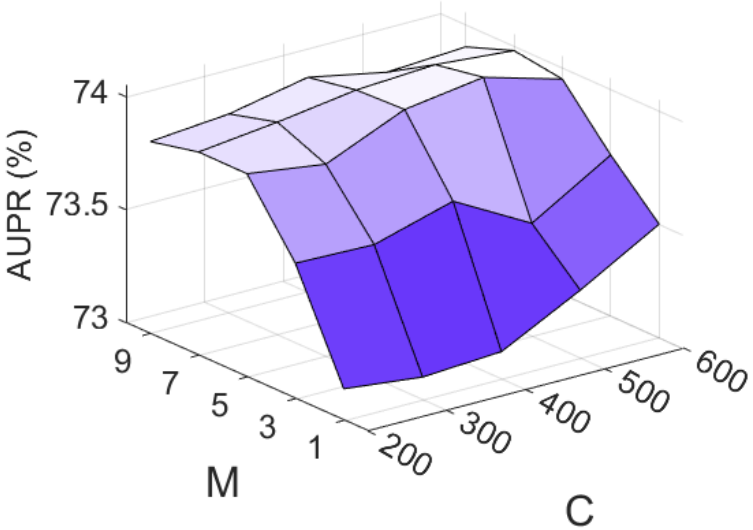}} 
\hspace{3mm}
\subfigure[Task 3]{\includegraphics[width=2.15in,height=1.4in]{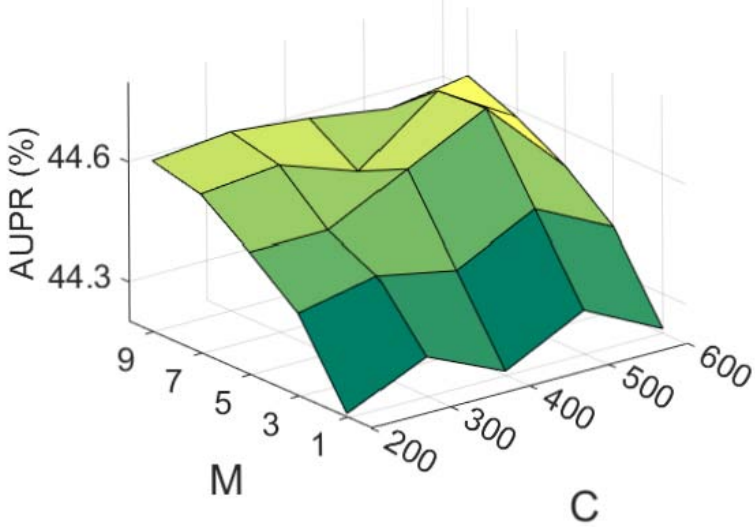}}
\caption{AUPR (\%) versus the clusters’ number $C$ and heads’ number $M$ on different tasks of Dataset 1.}
\vspace{-0.3cm}
\label{cmtask}
\end{figure*}

\begin{figure*}[t]
\centering 
\subfigure[Task 1]{\includegraphics[width=2.25in,height=1.3in]{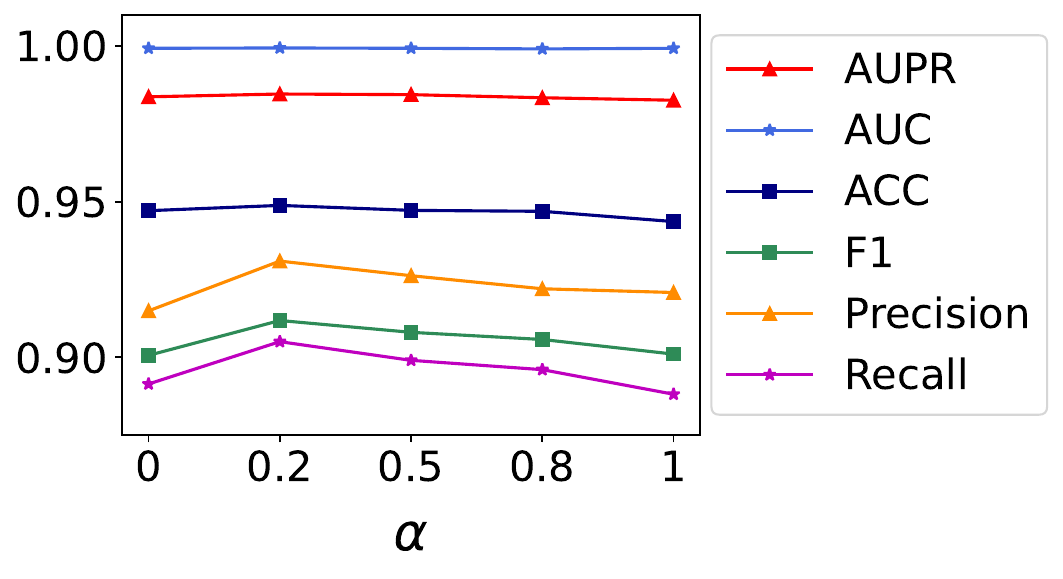}} 
\hspace{2mm}
\subfigure[Task 2]{\includegraphics[width=2.25in,height=1.3in]{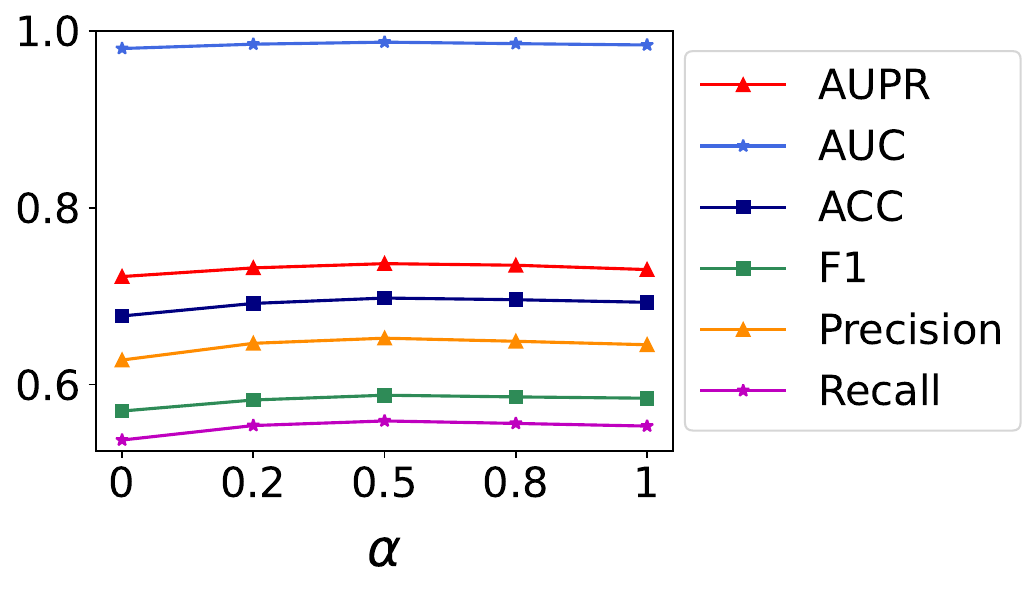}} 
\hspace{2mm}
\subfigure[Task 3]{\includegraphics[width=2.25in,height=1.3in]{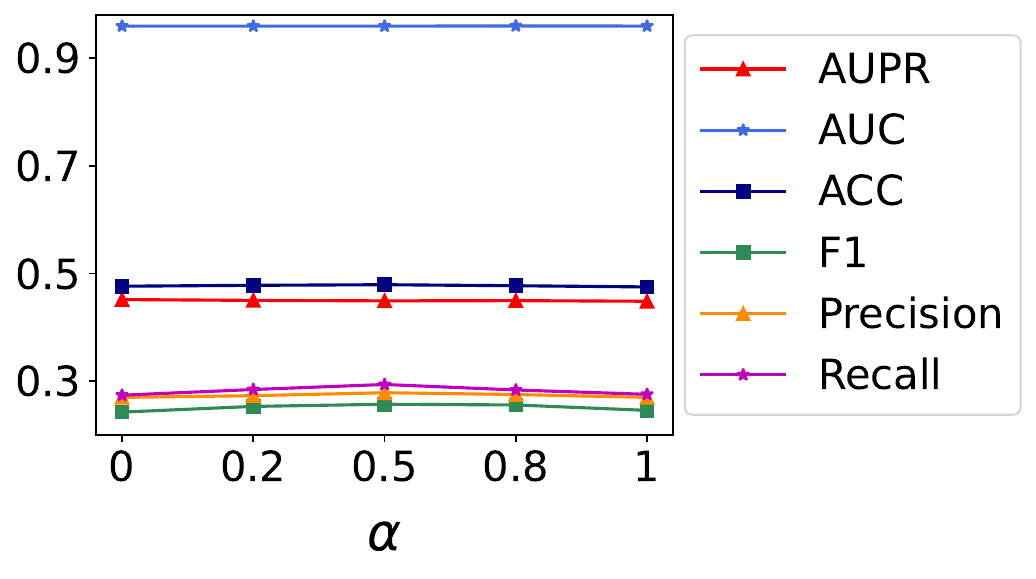}}
\vspace{-0.2cm}
\caption{Six metrics versus the regularization item weight $\alpha$ on different tasks of Dataset 1.}
\label{alpha3task}
\vspace{-1mm}
\end{figure*}

\subsection{Hyperparameter Exploring and Setting}\label{Hyperparameter}
The proposed HMGRL involves several hyperparameters, including batch size ($bs$), learning rate ($lr$), dropout rate ($dr$), training epochs ($te$), the dimension of drug embedding ($d'$), the embedding propagation distance ($L$), the output dimension of Encoder 1 ($d^{emb}$), the output dimension of CNN and Encoders 2 to 4 ($d^{att}$), the number of heads in the DSC ($M$), the number of clusters in DSC ($C$), and the weight of regularization item ($\alpha$).
To thoroughly explore the effects of each hyperparameter on prediction results, we employed a grid search strategy while keeping other parameters fixed. The optimal hyperparameter settings for the proposed HMGRL are listed in Table~\ref{parasearch}.
Among these hyperparameters, $L$, $d^{emb}$, $d^{att}$, $M$, $c$ and $\alpha$ are specific to the proposed HMGRL. 
Therefore, we will carefully analyze the influence of these six hyperparameters on the prediction performance of HMGRL and analyze the reasons.

\subsubsection{Impact of Embedding Propagation Distance $L$}
\label{maximumL}
The hyperparameter $L$ determines the maximum distance of embedding propagation and is pivotal in generating efficient RaGSEs for new drugs. 
To comprehend the impact of $L$ on the prediction performance, we conduct experiments on three tasks of Dataset 1, evaluating six metrics over varying $L$ values.
Referring to Figure \ref{n-hop} (a), in Task 1, HMGRL's metrics show a slight decrease as the parameter $L$ increases. This trend may be attributed to RaGSE diluting as $L$ increases. Despite this, HMGRL consistently demonstrates strong predictive performance.
On the other hand, for Tasks 2 and 3, the predictive capability of HMGRL displays heightened sensitivity to changes in $L$.
Specifically, HMGRL's performance dips to its nadir when $L$ is 0.
As $L$ increases, there's a marked enhancement in predictive performance.  
This is mainly attributed to the test DDIs of Tasks 2 and 3 containing new drugs.
These new drugs have no neighbors in the DDI graph and can not learn effective RaGSEs from the DDI graph.
Thus, the test data can not possess crucial information like the training data when $L$ is 0.
With an increased $L$, these new drugs can assimilate features from their connected drugs in the DDS graph, ensuring more effective RaGSEs. 
Moreover, an ideal propagation distance is essential for capturing effective contextual information. If the propagation distance is excessively long, it may lead to an undue dilution of information. In Tasks 2 and 3, when $L$ reaches 3, HMGRL achieves a balance in extracting local and global features, resulting in the highest metrics.
Based on our findings, we set $L$ as 0 in Task 1 and 3 in Tasks 2 and 3 to achieve the best predictive outcomes.

\subsubsection{Impact of dimensions $d^{att}$ and $d^{emb}$}
\label{twodimension}
The output dimensions of Encoder 1 and Encoders 2-4, represented as $d^{att}$ and $d^{emb}$, respectively, are important for our HMGRL framework.
Increasing $d^{att}$ and $d^{emb}$ can somewhat enhance HMGRL's generalization capabilities. 
Nevertheless, with the inclusion of new drugs in the test DDIs for Tasks 2 and 3, the significance of these hyperparameters varies across tasks.
To investigate the influence of $d^{att}$ and $d^{emb}$ on HMGRL's prediction performance, we conduct experiments across the three tasks of Dataset 1 and examine the changes in AUPR as $d^{att}$ and $d^{emb}$ varies.
Fig.~\ref{dimtask} (a) exhibits the findings for Task 1. 
For this task, we varied $d^{att}$ in the range $[200, 400, \dots, 800]$, and $d^{emb}$ in the range $[500, 700, \dots, 1700]$.
It's evident that the AUPRs slowly increase as $d^{att}$ increases and significantly increase as $d^{emb}$ increases.
This observation confirms the importance of RaGSEs to the model.
Finally, we set $d^{att}=200$ and $d^{emb}=1500$ for Task 1 to attain peak performance.
Figs.~\ref{dimtask} (b) and (c) display the comparisons on the Tasks 2 and 3, respectively.
In these two tasks, we set $d^{att}$ and $d^{emb}$ in the range $[200, 400, \dots, 1000]$.
We observe that the increase of both $d^{att}$ and $d^{emb}$ have obviously positive impacts on AUPRs.
This indicates the effectiveness of drugs' RaGSEs and multiple biochemical attributes.
To balance accuracy and efficiency, we set $d^{att}=800$ and $ d^{emb}=800$ for Tasks 2 and 3.

\begin{figure*}[t]
\centering
\subfigure[Task 1]{\includegraphics[width=2.25in,height=1.2in]{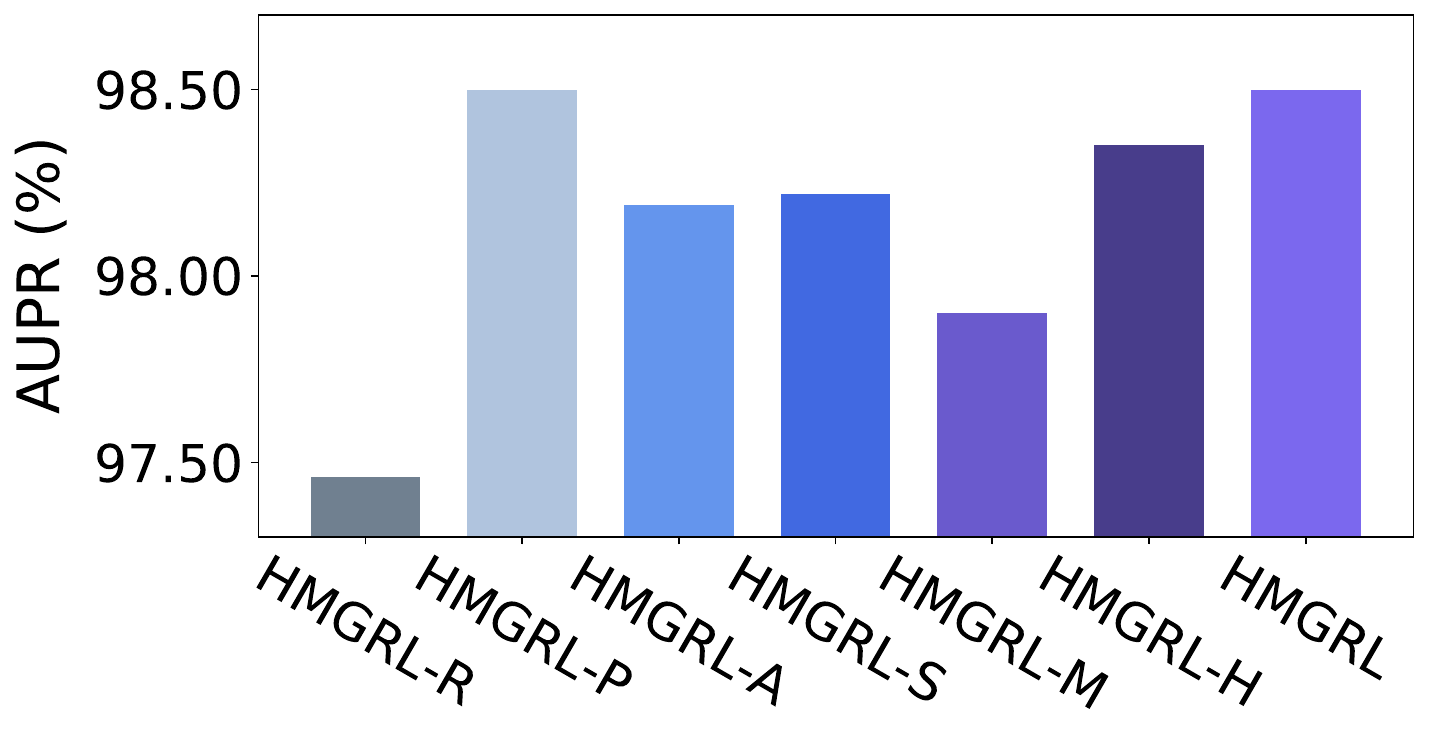}} 
\hspace{1mm}
\subfigure[Task 1]{\includegraphics[width=2.25in,height=1.2in]{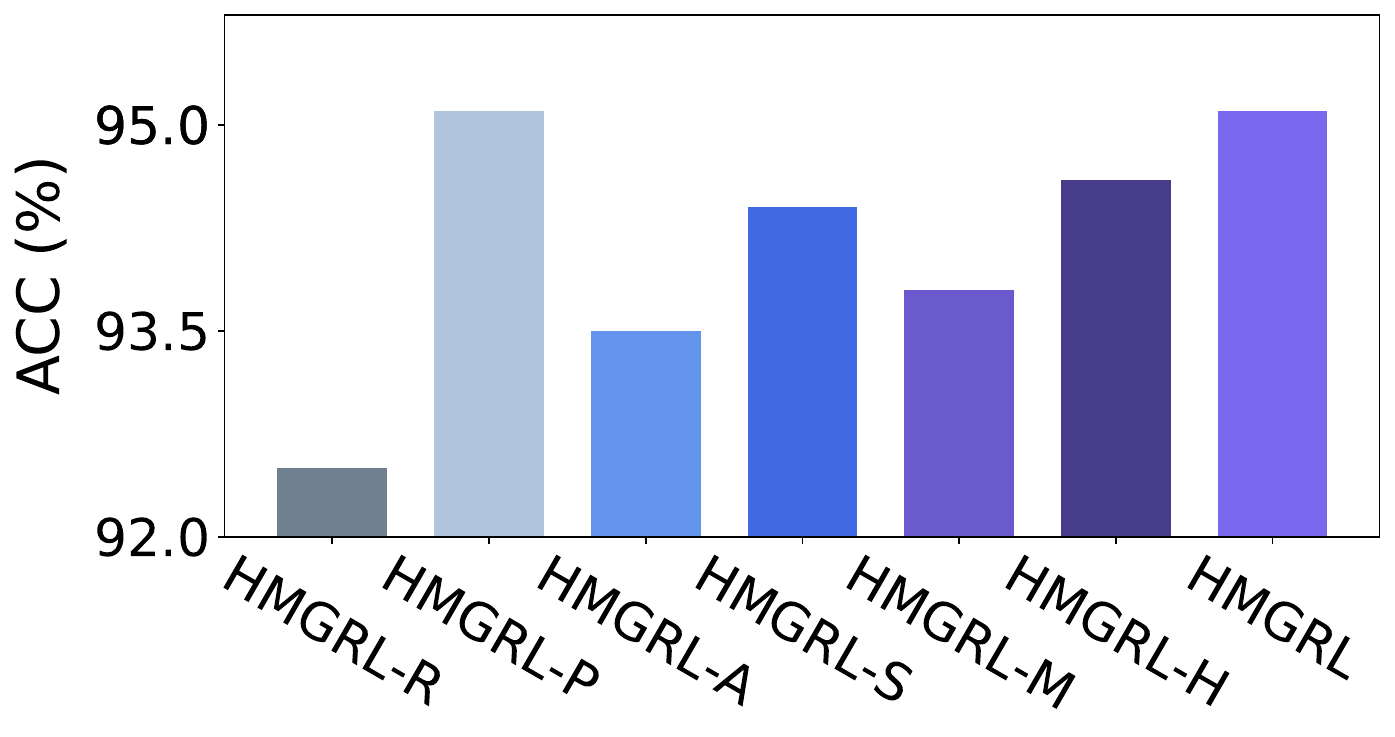}} 
\hspace{1mm}
\subfigure[Task 1]{\includegraphics[width=2.25in,height=1.2in]{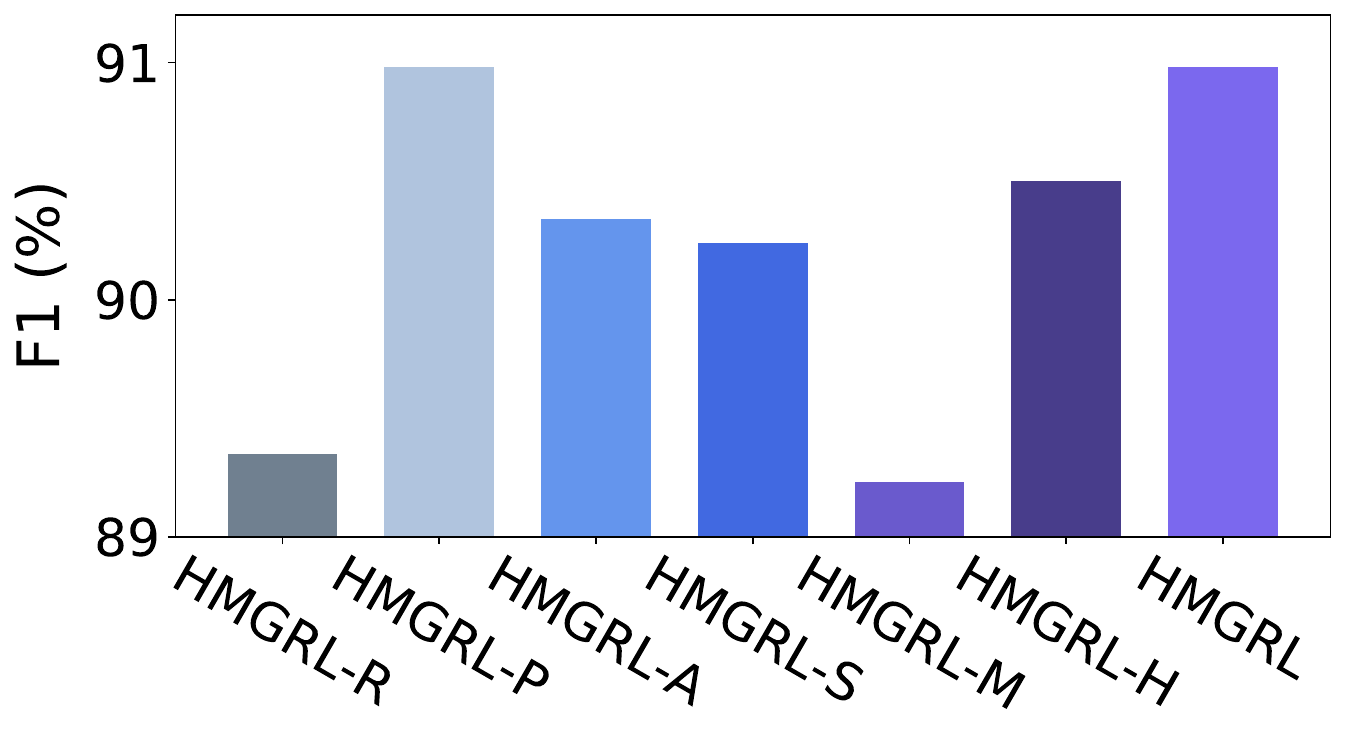}}\\
\vspace{-0.3cm}
\subfigure[Task 2]{\includegraphics[width=2.25in,height=1.2in]{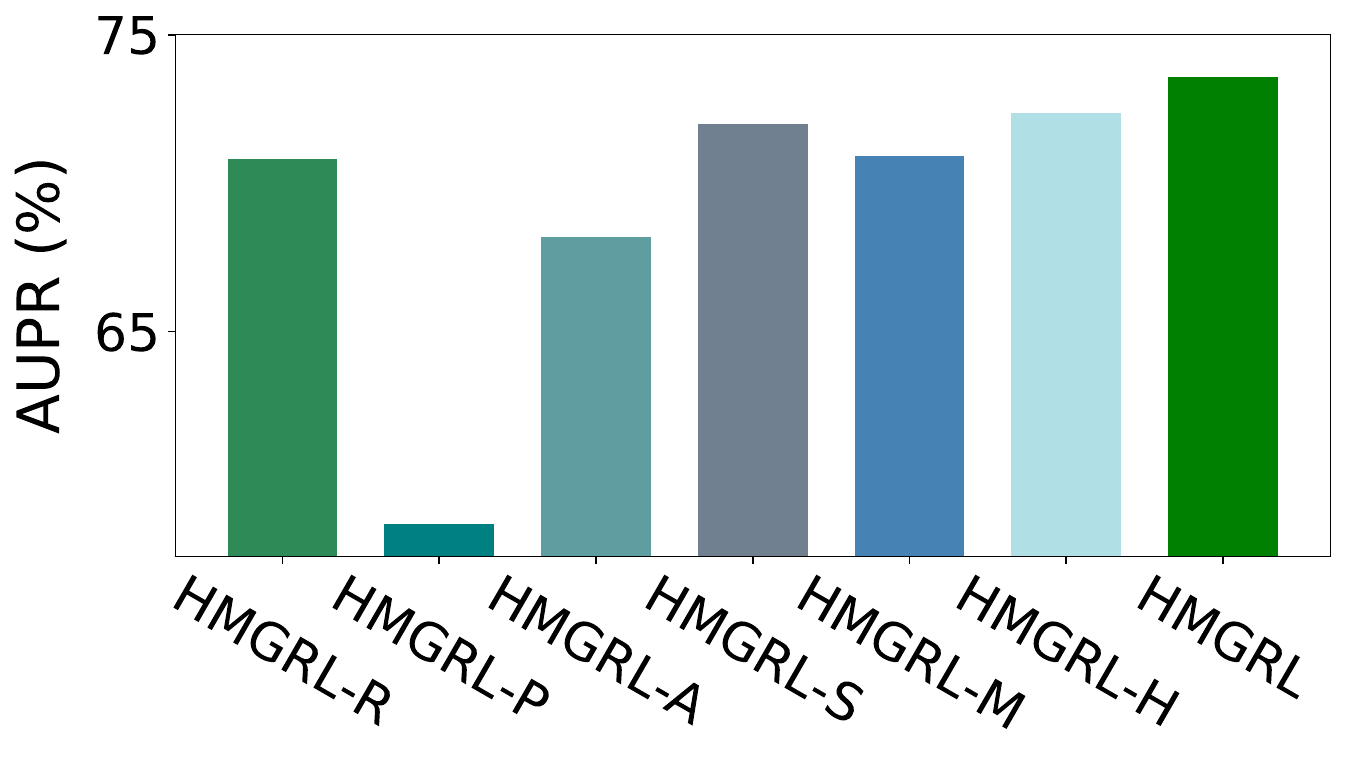}} 
\hspace{1mm}
\subfigure[Task 2]{\includegraphics[width=2.25in,height=1.2in]{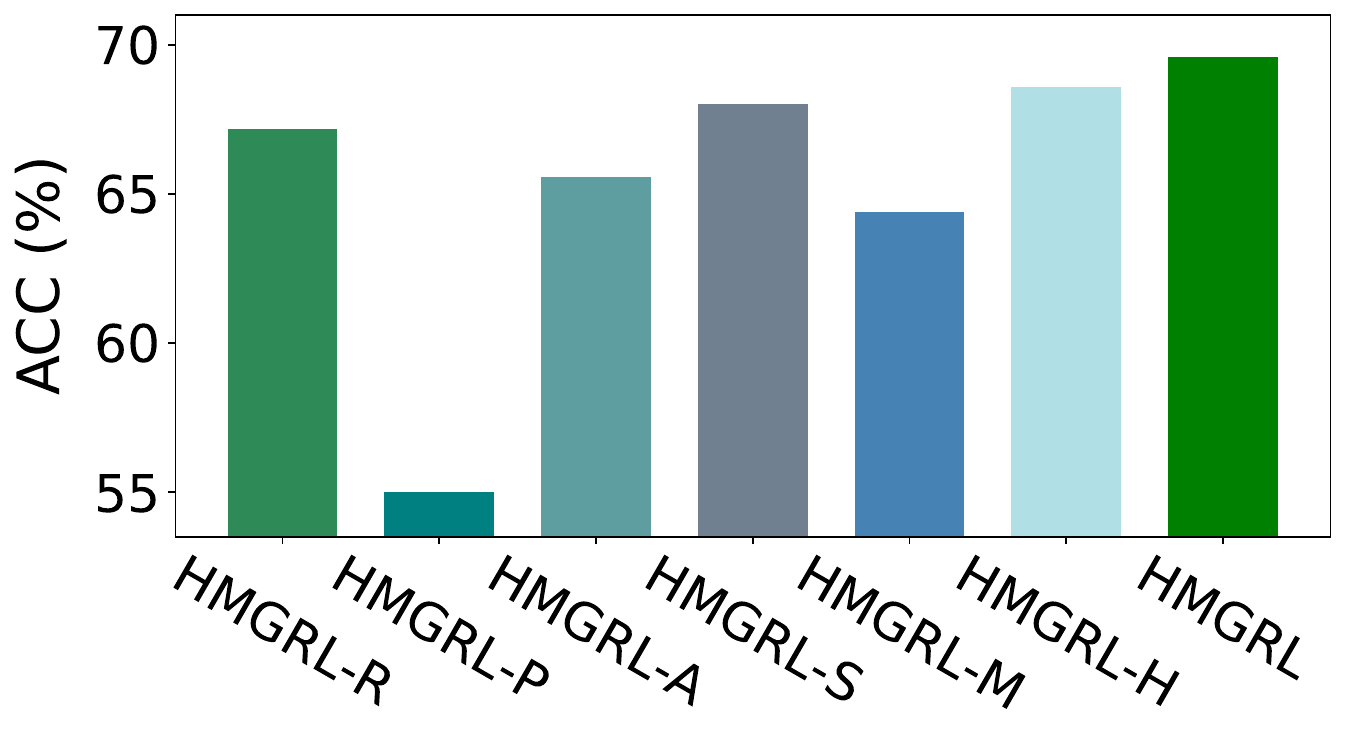}}
\hspace{1mm}
\subfigure[Task 2]{\includegraphics[width=2.25in,height=1.2in]{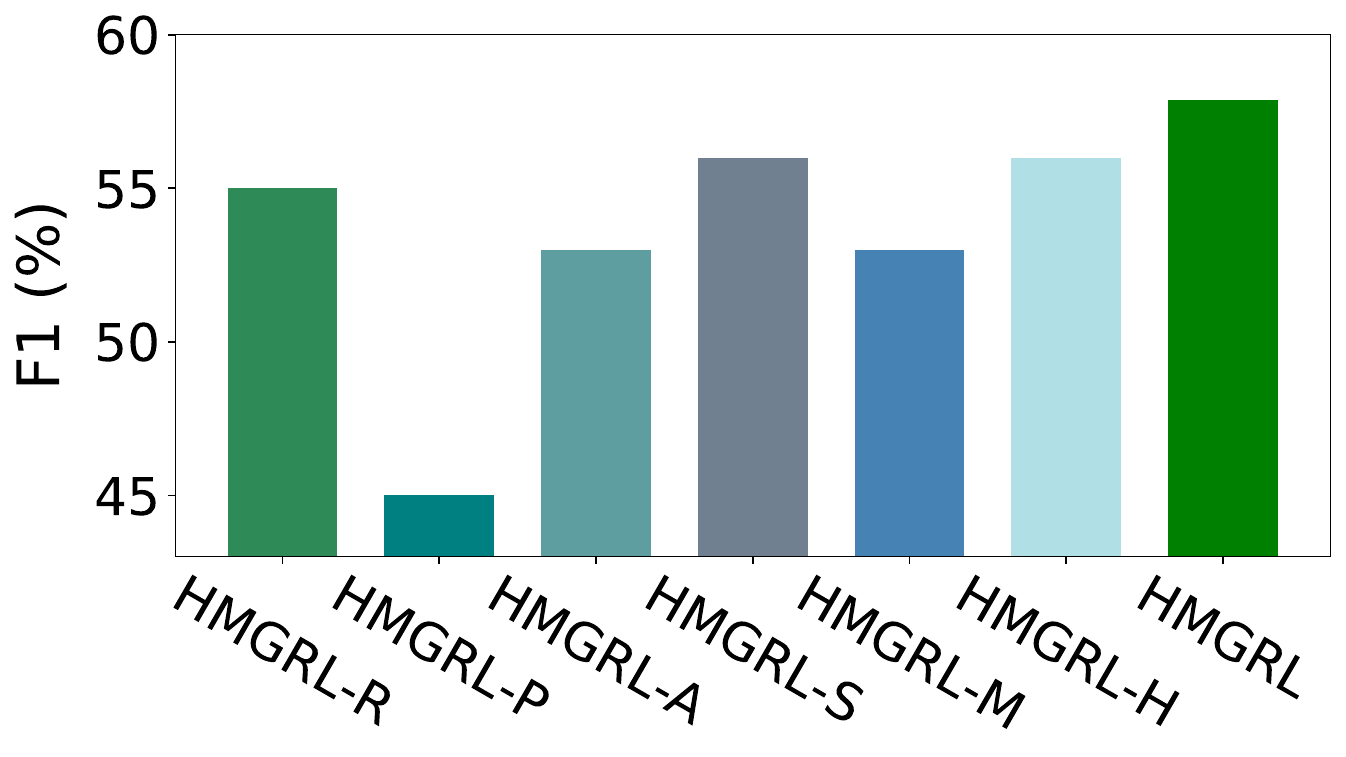}}\\ 
\vspace{-0.3cm}
\subfigure[Task 3]{\includegraphics[width=2.25in,height=1.2in]{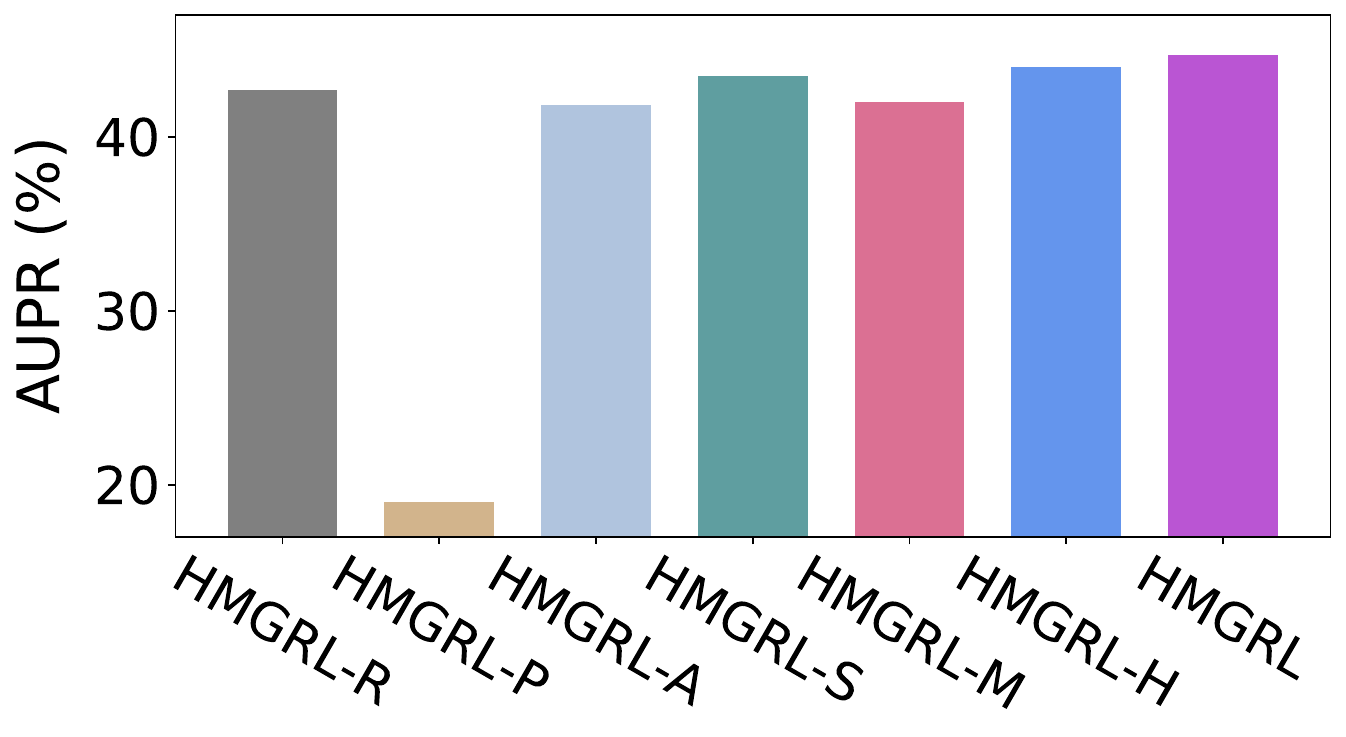}}
\hspace{1mm}
\subfigure[Task 3]{\includegraphics[width=2.25in,height=1.2in]{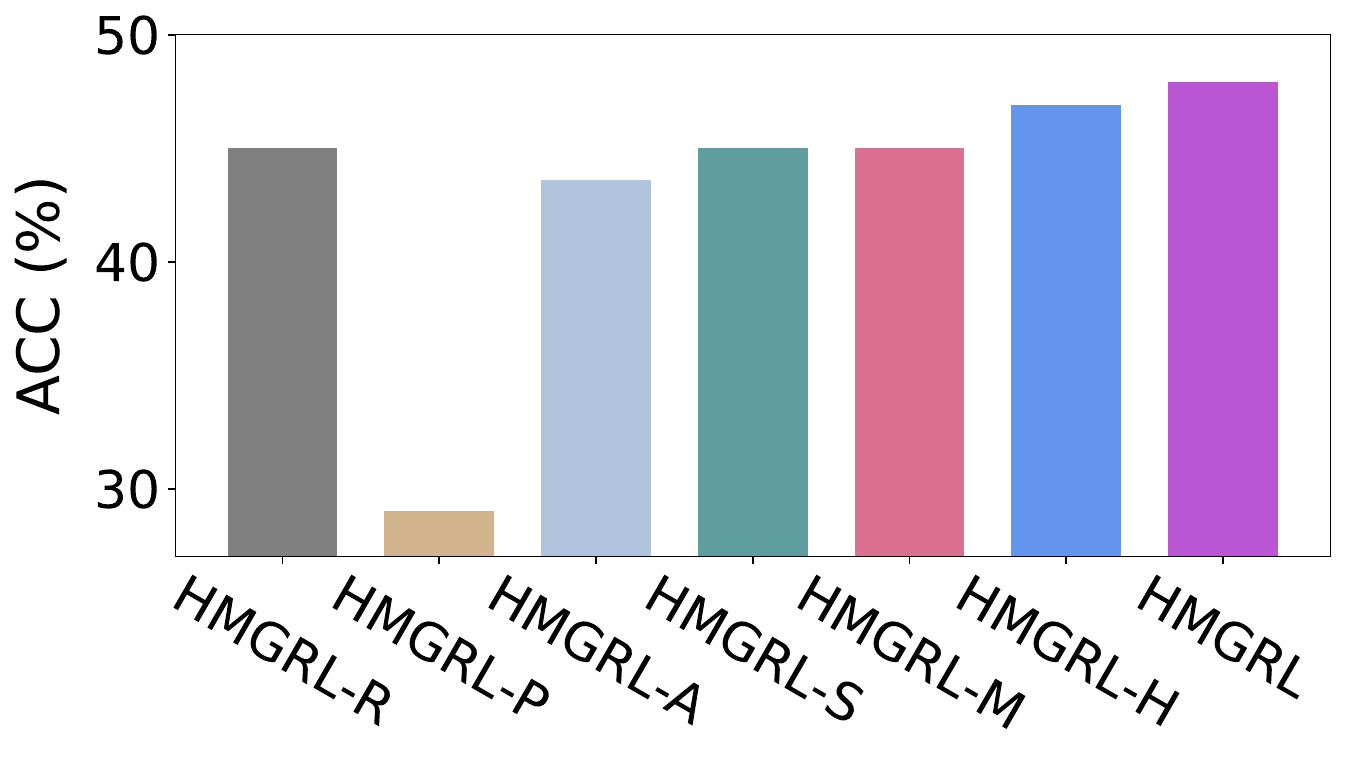}} 
\hspace{1mm}
\subfigure[Task 3]{\includegraphics[width=2.2in,height=1.2in]{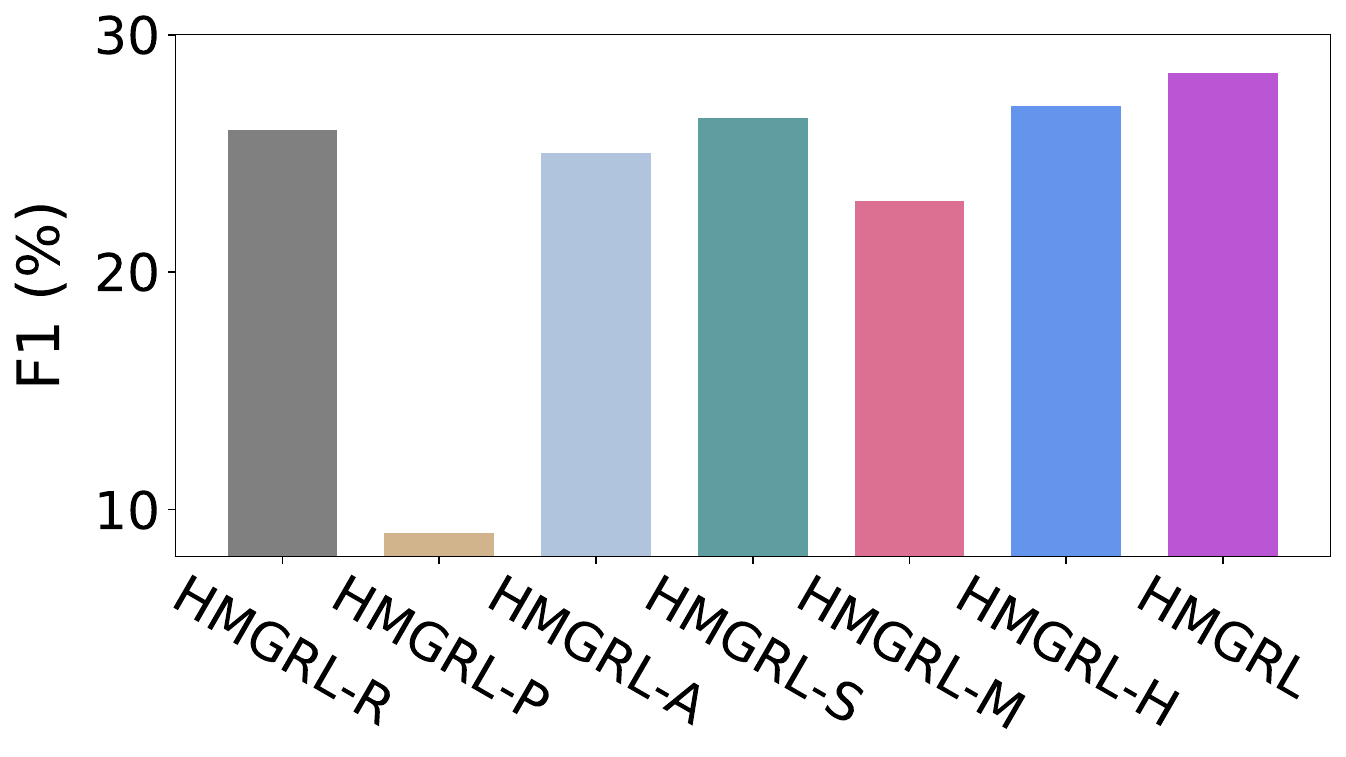}}
\caption{Experimental results of HMGRL and its six variants in terms of AUPR, AUC, and F1 on different tasks of Dataset 1. (a)-(c) Task 1. (d)-(f) Task 2. (g)-(i) Task 3.}
\label{variant123}
\vspace{-0.4cm}
\end{figure*}
 
\subsubsection{Impact of Clusters' Number $C$ and Heads' Number $M$}
\label{clustershead}
In the original SC \cite{SC0}, clusters' number $C$ impacts clustering performance.
Meanwhile, the multi-head strategy can stabilize the learning process of self-attention within DSC.
To get the best out of DSC, we conduct experiments to evaluate prediction performance under various configurations of $C$ and $M$. 
Fig.~\ref{cmtask} shows the values of AUPR changing with the variations in $C$ and $M$ on three tasks of Dataset 1.
Task 1 has a batch size of 512, while Tasks 2 and 3 have a batch size of 1024.
Consequently, $C$ ranges from $100$ to $400$ within Task 1 and from $200$ to $600$ within Tasks 2 and 3.
we set $M$ in the range $[1, 3, \dots, 9]$.
As Fig.~\ref{cmtask} shows, the model's performance remains relatively stable as $C$ varies, demonstrating the robustness of our model.
Meanwhile, $M$ has a relatively obvious impact on AUPRs.
As $M$ increases, AUPRs slowly grow.
This phenomenon confirms the positive role of the multi-head strategy in the learning process of self-attention.
Finally, we set $M=5$ for Tasks 1 to 3.
$C=200$ for Tasks 1, $C=400$ for Tasks 2 and 3.

\subsubsection{Impact of Regularization Item Weight $\alpha$}
\label{alpha}
The hyperparameter $\alpha$ represents the weight of the regularization item $\mathcal{L}_{DSC}$, thus determining the influence of the graph-cutting strategy on model training.
To explore the impact of the graph-cutting strategy on the prediction performance, we carried out experiments on three tasks of Dataset 1, evaluating six metrics over varying $\alpha$ values.
Fig.~\ref{alpha3task} shows the experimental results. 
As can be seen, although the $\alpha$ value has varying impacts on the model's performance in different tasks, our model's performance remains stable, proving our model's stability.
Within the range of values for $\alpha$, the graph-cutting strategy has a positive impact on our HMGRL most of the time, which also proves the effectiveness of our graph-cutting strategy.
Finally, to get the best prediction performance, 
we set $\alpha=0.2$ for Task 1, $\alpha=0.5$ for Tasks 2 and 3.

\subsection{Ablation Study}\label{Variants}
Our HMGRL initially learns RaGSEs for known drugs and propagates RaGSEs for improving new drugs' embedding.
Next, HMGRL incorporates drugs’ RaGSEs and multiple biochemical attributes to learn comprehensive DP features. 
Subsequently, HMGRL constructs the MVDSC module to capture underlying correlations between DPs from distinct perspectives. 
To demonstrate the effectiveness and necessity of each improved component, we perform further comparisons between our HMGRL and its six variants:

$\bullet~$HMGRL-R: A variant of HMGRL that does not generate drug RaGSEs. In this case, we directly use the drug's multiple biochemical attributes to compute the DP features during multi-source feature learning: $\widetilde{\bf h}_{k} = || \left(\widetilde{\bf h}_{smi,k}, \widetilde{\bf h}_{tar,k},\widetilde{\bf h}_{enz,k},\widetilde{\bf h}_{sub,k}\right)$.  

$\bullet~$HMGRL-P: A variant of HMGRL that does not improve embeddings for new drugs by RaGSE propagation. In this case, we directly use the output of
Eq. (\ref{rgcn}) as the input of Encoder 1.

$\bullet~$HMGRL-A: A variant of HMGRL that omits these three attributes (targets, enzymes, and molecular substructures) during multi-source feature learning. In this version, the DP features are presented as: $\widetilde{\bf h}_{k} = || \left(\widetilde{\bf h}_{smi,k},\widetilde{\bf h}_{emb,k}\right)$.

$\bullet~$HMGRL-S: A variant of HMGRL that does not consider the drug's SMILES strings during multi-source feature learning. In this version, the DP features are presented as: $\widetilde{\bf h}_{k} = || \left(\widetilde{\bf h}_{emb,k},\widetilde{\bf h}_{tar,k},\widetilde{\bf h}_{enz,k},\widetilde{\bf h}_{sub,k}\right)$.

$\bullet~$HMGRL-M: A variant of HMGRL omits the MVDSC module. The DP features $\widetilde{{\bf H}}$ is directly used for DDI prediction. The formula (\ref{predict}) is represented as ${\bf Y}=\operatorname{Decoder}\left(\widetilde{{\bf H}};\Theta_{\mathrm{MLP}}\right)$.

$\bullet~$HMGRL-H: A variant of HMGRL that only focuses on one implicit DP correlation, In this scenario, $\widetilde{\bf H}$ is the only input of DSCs 1 to 4.

Fig.~\ref{variant123} presents the results of HMGRL and its six variants in terms of AUPR, Accuracy, and F1 on Tasks 1 to 3 of Dataset 1.
Fig.~\ref{variant123}(a)-(c) shows the results on Task 1, Fig.~\ref{variant123}(d)-(f) exhibits for Task 2, and Fig.~\ref{variant123}(g)-(i) exhibits for Task 3.
Obviously, these metrics for HMGRL are higher than those for six variants, indicating the effectiveness of considering RaGSE learning and propagation, multi-source feature learning, and multiple underlying associations between DPs.

% % \\tiny\\scriptsize\\footnotesize\\small\\normalsize\\large\\Large\\LARGE\\huge
\begin{table}[t]\footnotesize
\caption{Prediction performances (\%) of different methods on different tasks of Dataset 1. The best results are in bold.}
\vspace{-0.2cm}
\tabcolsep=2pt%% 
\begin{tabular}{lccccccc}
\toprule
& Method $~$ &  AUPR $~$ & AUC$~$ & ACC $~$ & F1$~$ & Precision$~$ & Recall \\
\midrule
\multirow{7}{*}{Task$~$1}
& DDIMDL          & 92.08 & 99.76 & 88.52 & 75.85 & 84.71 & 71.82 \\
& RANEDDI         & 96.57 & 99.80 & 92.28 & 87.17 & 87.47 & 87.01 \\
& MDF-SA-DDI      & 97.37 & 99.89 & 93.01 & 88.78 & 90.85 & 87.60 \\
& MAEDDIE         & 97.76 & 99.89 & 94.26 & 88.65 & 90.70 & 87.81 \\
& MCFF-MTDDI      & 97.57 & 99.85 & 93.50 & 89.18 & 91.00 & 88.20 \\ 
& MM-GANN-DDI     & 97.86 & 99.89 & 93.86 & 89.80 & 90.88 & 89.50 \\ 
& \bf{HMGRL}      & \bf{98.46} & \bf{99.94} & \bf{94.88} & \bf{91.18} & \bf{93.09} & \bf{90.50}   \\  
\midrule
\multirow{5}{*}{Task$~$2}
& DDIMDL          & 65.58 & 97.99 & 64.15 & 44.60 & 56.07 & 43.19 \\
& MDF-SA-DDI      & 67.76 & 94.97 & 66.33 & 55.84 & 65.47 & 50.78 \\
& MAEDDIE         & 68.93 & 96.81 & 67.31 & 55.65 & 65.52 & 50.51  \\
& MCFF-MTDDI      & 68.00 & 95.00 & 66.50 & 55.74 & \bf{65.61} & 51.39 \\ 
& MM-GANN-DDI     & 68.55 & 95.32 & 67.05 & 55.80 & 65.18 & 51.56 \\ 
& \bf{HMGRL}      & \bf{73.37} & \bf{98.52} &\bf{69.58} &\bf{58.59} & 65.04 & \bf{55.68}   \\ 
\midrule
\multirow{5}{*}{Task$~$3}
& DDIMDL          & 36.35 & 95.12 & 40.75 & 15.90 & 24.08 & 14.52 \\
& MDF-SA-DDI      & 38.73 & 86.30 & 43.38 & 23.29 & 27.15 & 22.26  \\
& MAEDDIE         & 39.62 & 92.71 & 45.10 & 24.05 & 27.52 & 22.51   \\
& MCFF-MTDDI      & 38.70 & 87.01 & 44.00 & 24.37 & \bf{28.23} & 23.51  \\ 
& MM-GANN-DDI     & 37.86 & 87.89 & 43.86 & 25.05 & 27.74 & 24.80  \\ 
&\bf{HMGRL}       & \bf{44.71} & \bf{95.93} & \bf{47.91} & \bf{26.71} & 27.85 & \bf{28.38}  \\ 
\bottomrule
\end{tabular}
\vspace{-0.4cm}
\label{dataset1}
\end{table}

\begin{table}[t]\footnotesize
\caption{Prediction performances (\%) of different methods on different tasks of Dataset 2. The best results are in bold.}
\vspace{-0.2cm}
\tabcolsep=2pt%%
\begin{tabular}{lccccccc}
\toprule
& Method $~$ &  AUPR $~$ & AUC$~$ & ACC $~$ & F1$~$ & Precision$~$ & Recall \\
\midrule
\multirow{7}{*}{Task$~$1}
& DDIMDL     & 88.24 & 98.92 & 84.01  & 78.00 & 76.78 & 85.80 \\
& RANEDDI    & 92.25 & 98.72 & 86.11  & 81.10 & 81.55 & 90.84 \\
& MDF-SA-DDI & 93.85 & 99.79 & 87.25  & 82.20 & 75.18 & 91.98 \\
& MAEDDIE    & 94.26 & 99.85 & 88.55  & 84.64 & 81.20 & 91.21\\
& MCFF-MTDDI & 95.32 & 99.84 & 90.10  & 86.31 & 83.00 & 91.22 \\ 
& MM-GANN-DDI& 95.76 & 99.89 & 90.66  & 87.80 & 85.62 & 91.50  \\ 
& \bf{HMGRL} & \bf{98.36} & \bf{99.97} & \bf{93.92}  & \bf{92.09} & \bf{92.07} & \bf{92.71}\\ 
\midrule
\multirow{5}{*}{Task$~$2}
& DDIMDL     & 61.59 & 97.37 & 60.39 & 44.70 & 50.80 & 44.65 \\
& MDF-SA-DDI & 64.86 & 96.67 & 63.33 & 47.85 & 53.17 & 49.00 \\
& MAEDDIE    & 65.33 & 96 80 & 64 62 & 48 45 & 55.00 & 47.81 \\
& MCFF-MTDDI & 65.64 & 96.01 & 64.22 & 51.40 & 54.07 & 50.87 \\ 
& MM-GANN-DDI& 66.46 & 96.82 & 65.05 & 52.30 & 55.62 & 51.47  \\ 
& \bf{HMGRL} & \bf{73.35} & \bf{ 98.71} & \bf{67.68} & \bf{57.36} & \bf{56.70} & \bf{60.70}  \\ 
\midrule
\multirow{5}{*}{Task$~$3}
& DDIMDL     & 37.39 & 94.50 & 41.11 & 18.23 & 27.15 & 16.91 \\
& MDF-SA-DDI & 40.92 & 93.58 & 45.30 & 23.99 & 27.68 & 23.36 \\
& MAEDDIE    & 40.99 & 93.98 & 46.06 & 24.21 & 28.75 & 23.86 \\
& MCFF-MTDDI & 41.36 & 93.40 & 45.45 & 24.27 & 28.57 & 23.48 \\ 
& MM-GANN-DDI& 41.86 & 93.55 & 46.57 & 25.55 & 28.96 & 25.00  \\ 
& \bf{HMGRL} & \bf{48.05} & \bf{97.51} & \bf{49.68} & \bf{27.36} & \bf{29.70} & \bf{27.10}  \\  
\bottomrule
\vspace{-0.5cm}
\label{dataset222}
\end{tabular}
\end{table}

\begin{figure*}[t]
\centering
\subfigure{\includegraphics[width=3.3in,height=2.7in]{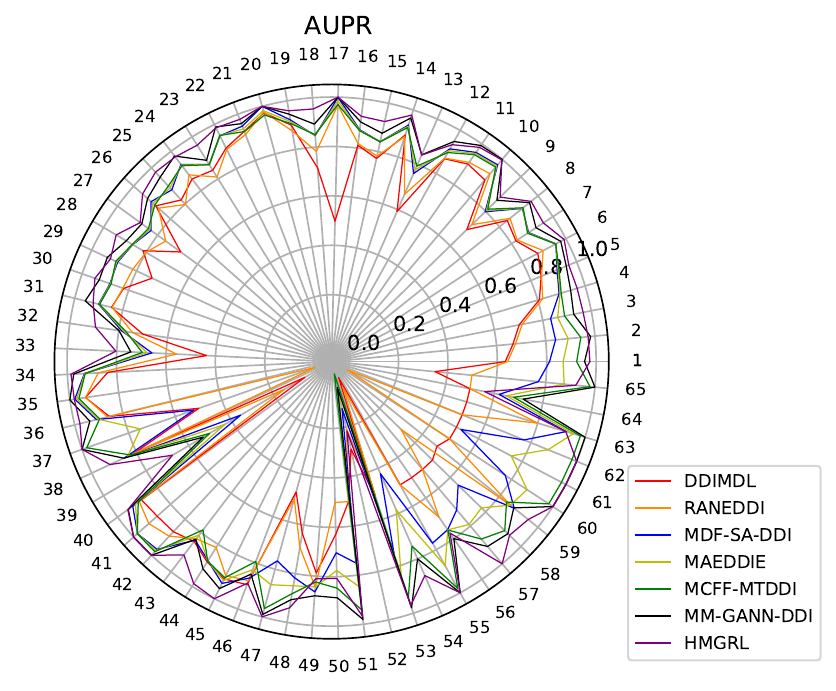}} 
\subfigure{\includegraphics[width=3.3in,height=2.7in]{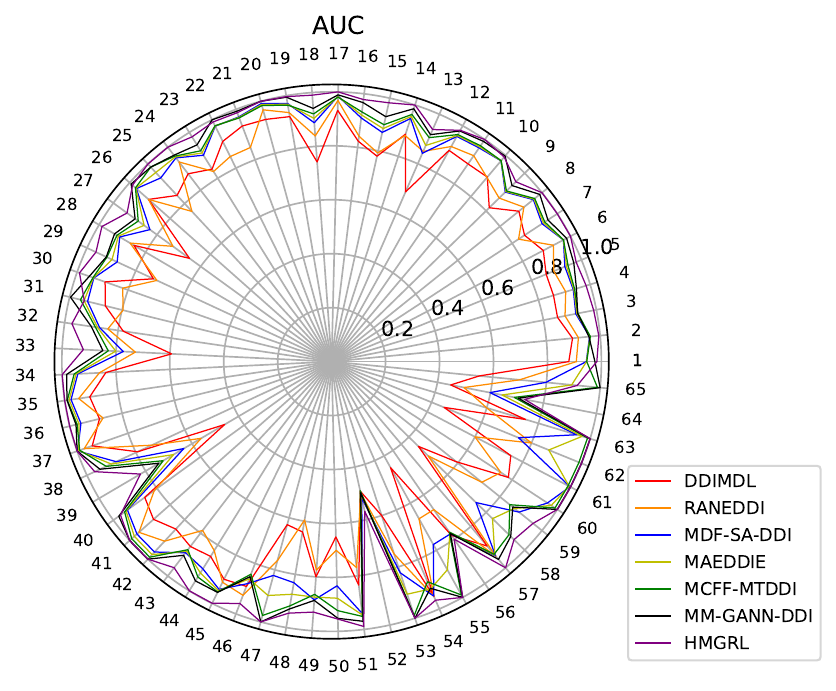}} 
\vspace{-0.2cm}
\caption{The AUPR and AUC scores of different prediction methods for each event on Task 1 of Dataset 1.}
\label{zzzz}
\end{figure*}

\begin{figure*}[t]
\centering
\subfigure[Raw Data]{\includegraphics[width=1.6in,height=1.1in]{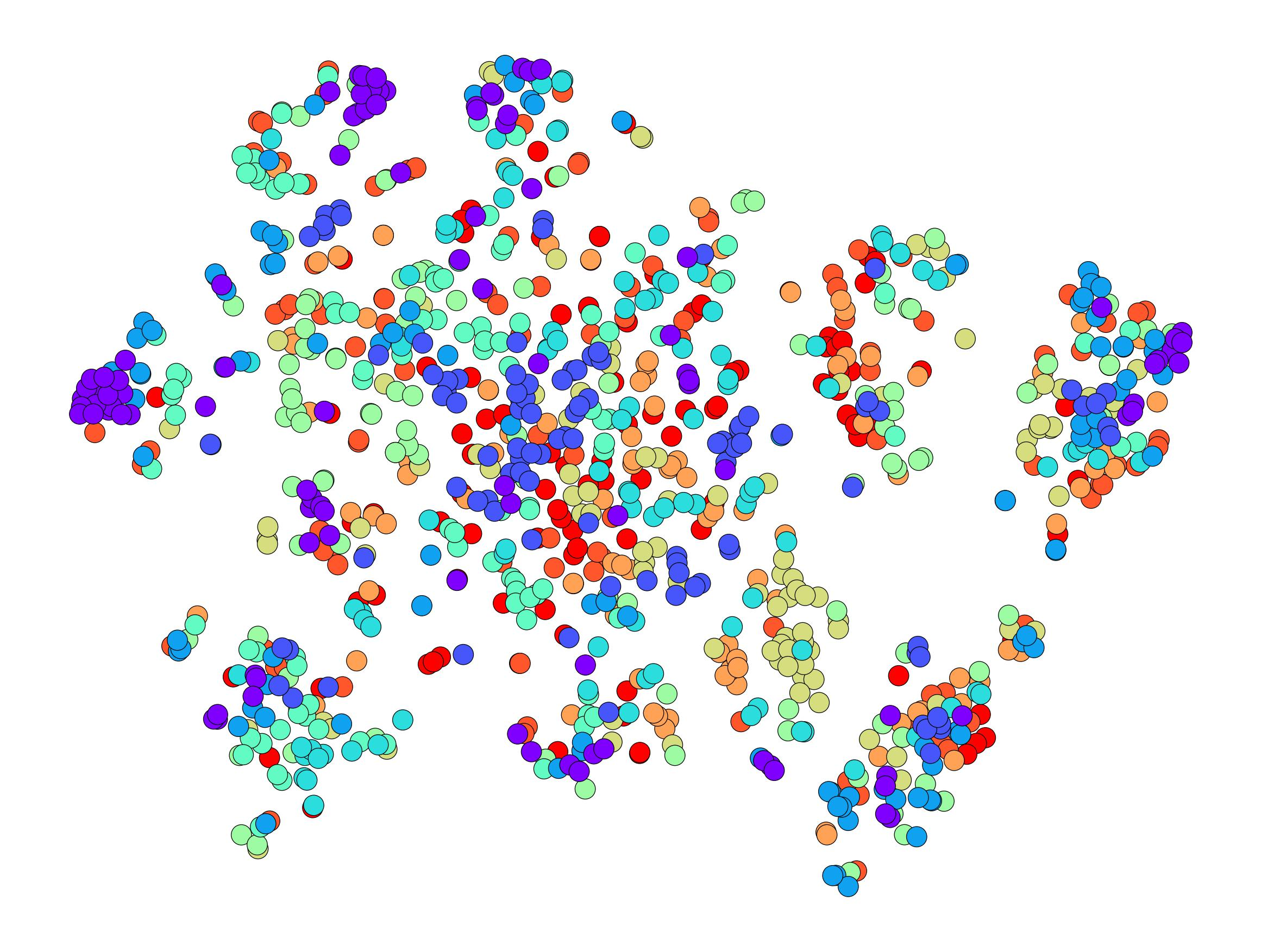}} 
\hspace{3mm}
\subfigure[DDIMDL]{\includegraphics[width=1.6in,height=1.1in]{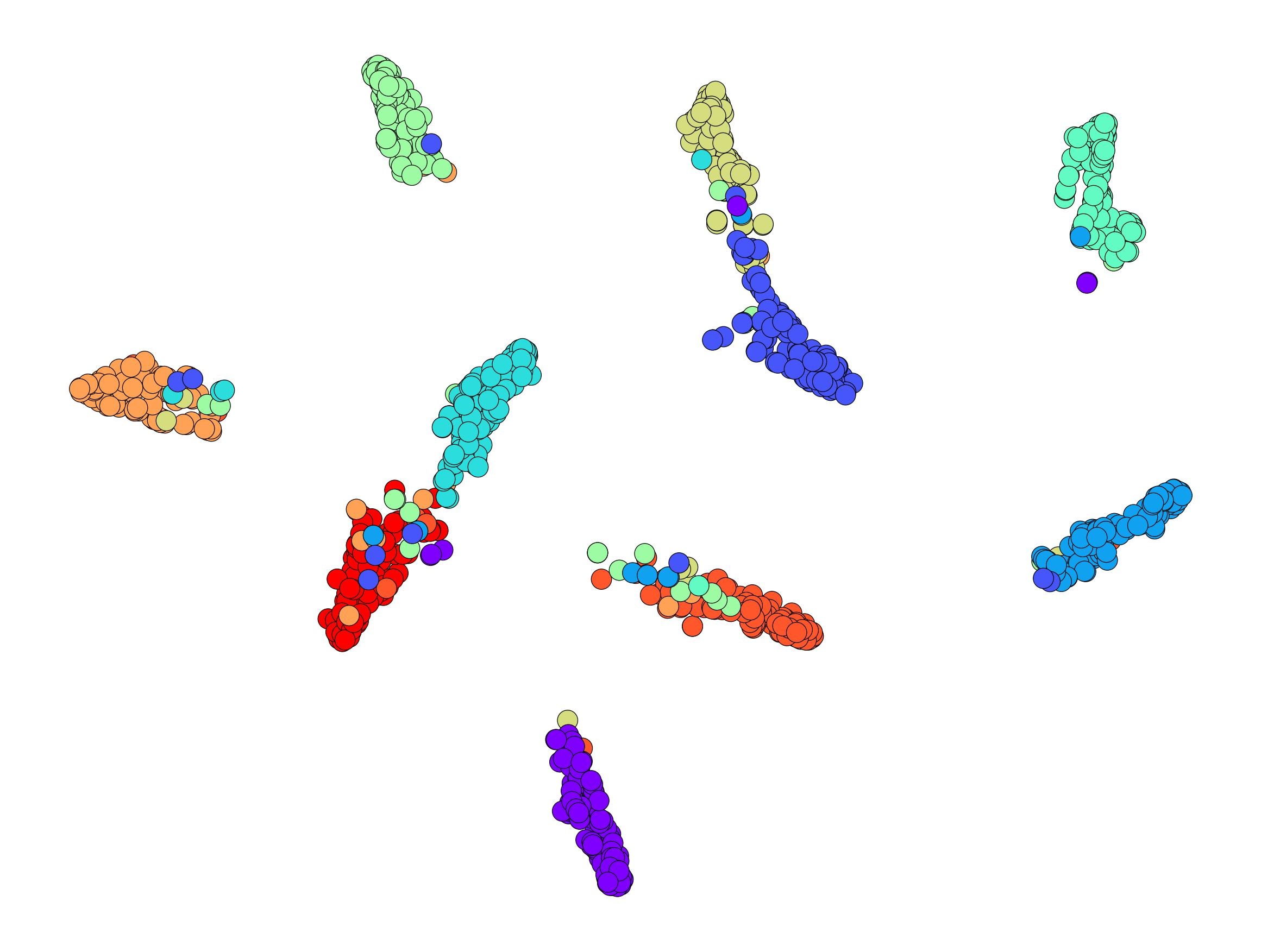}} 
\hspace{3mm}
\subfigure[RANEDDI]{\includegraphics[width=1.6in,height=1.1in]{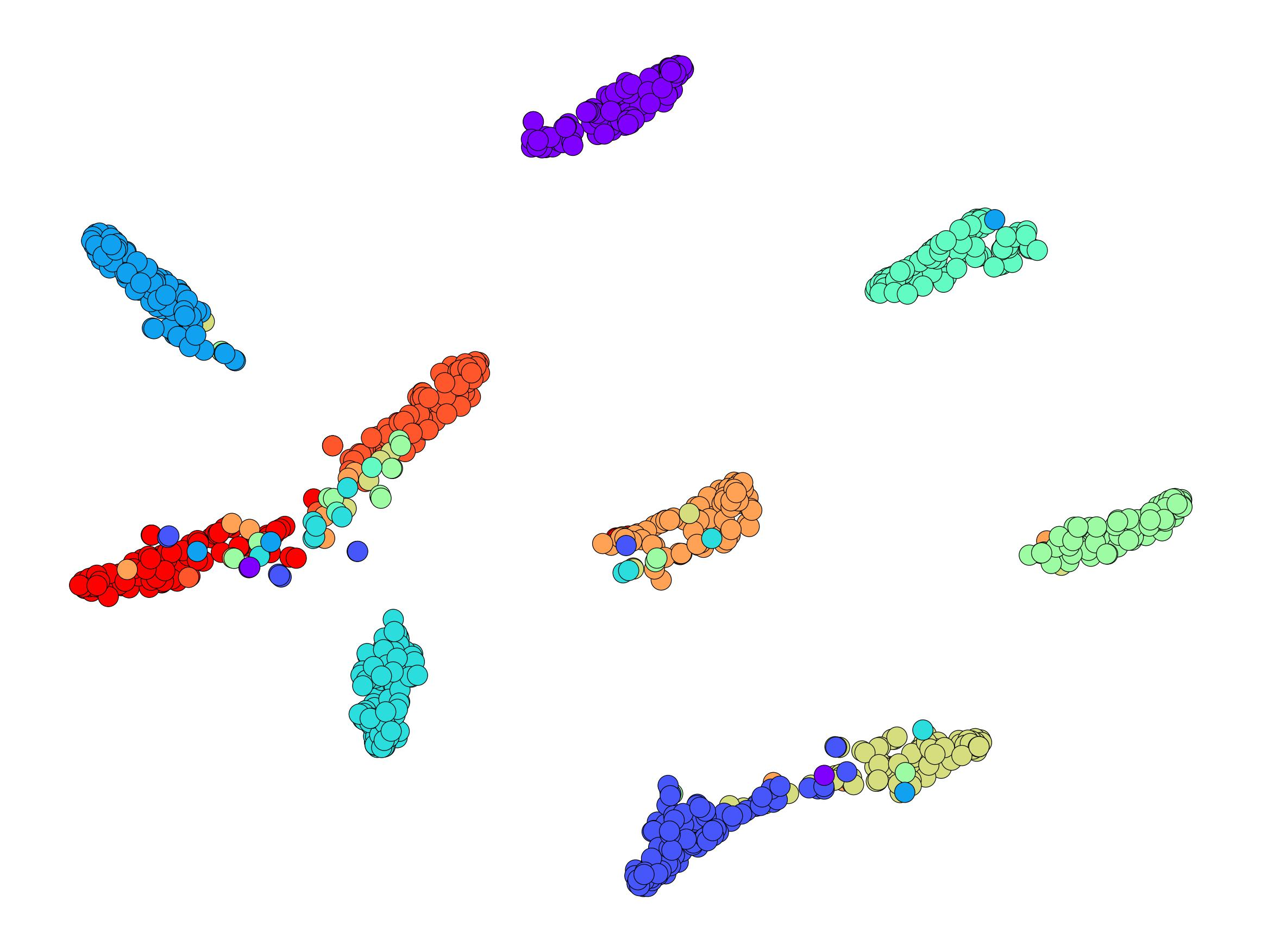}}
\hspace{3mm}
\subfigure[MDF-SA-DDI]{\includegraphics[width=1.6in,height=1.1in]{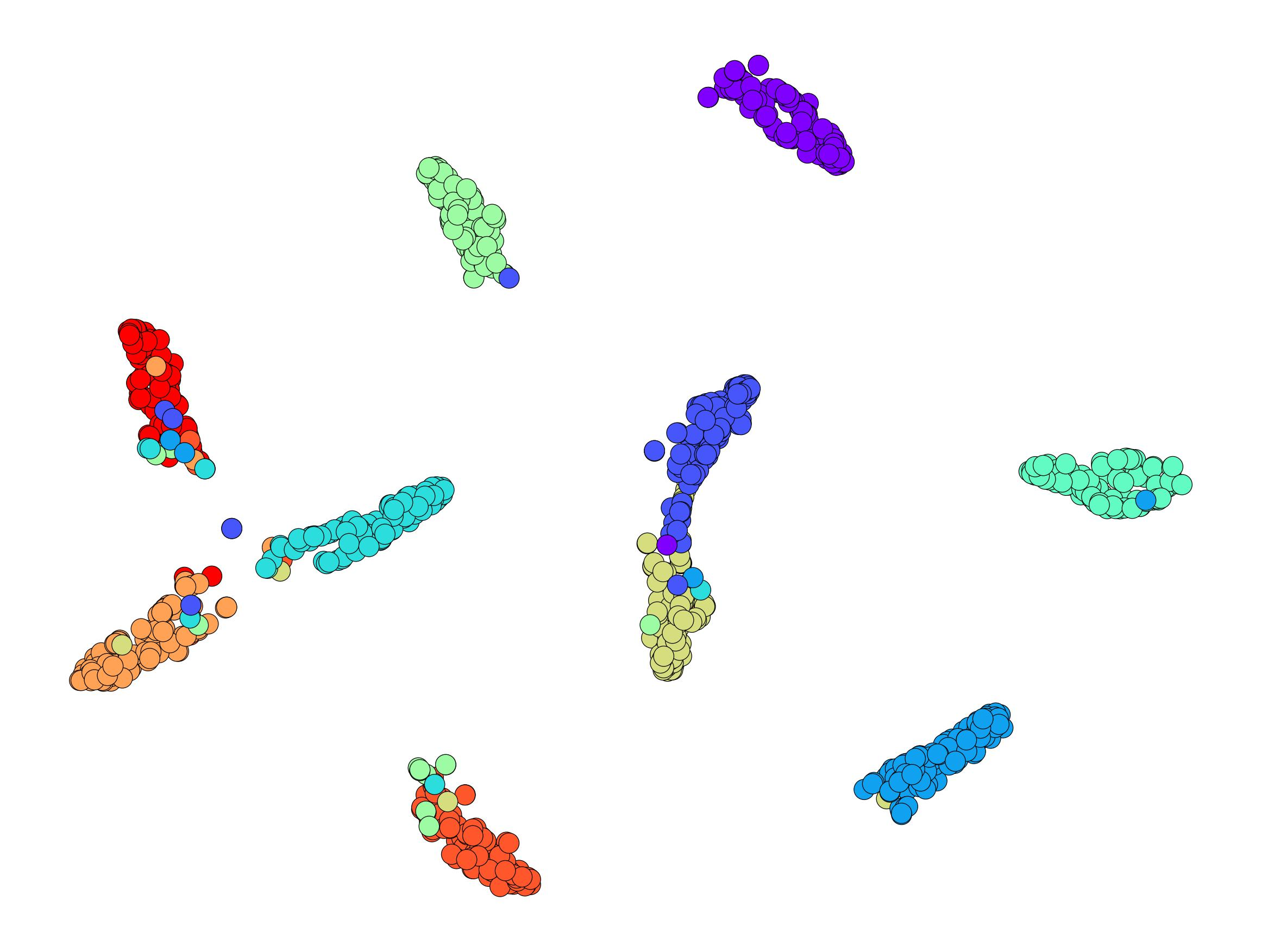}}\\
\subfigure[MAEDDIE]{\includegraphics[width=1.6in,height=1.1in]{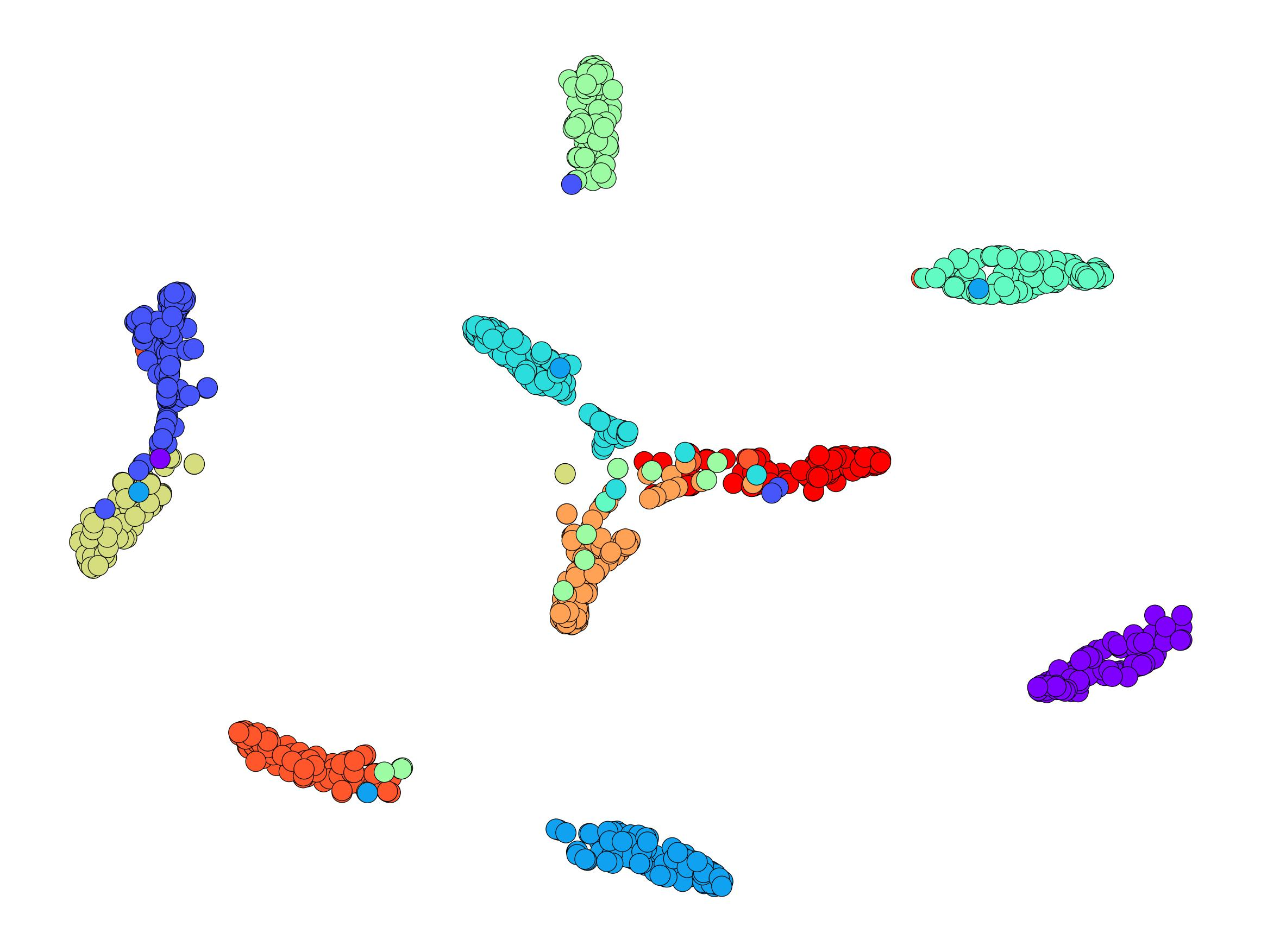}} 
\hspace{3mm}
\subfigure[MCFF-MTDDI]{\includegraphics[width=1.6in,height=1.1in]{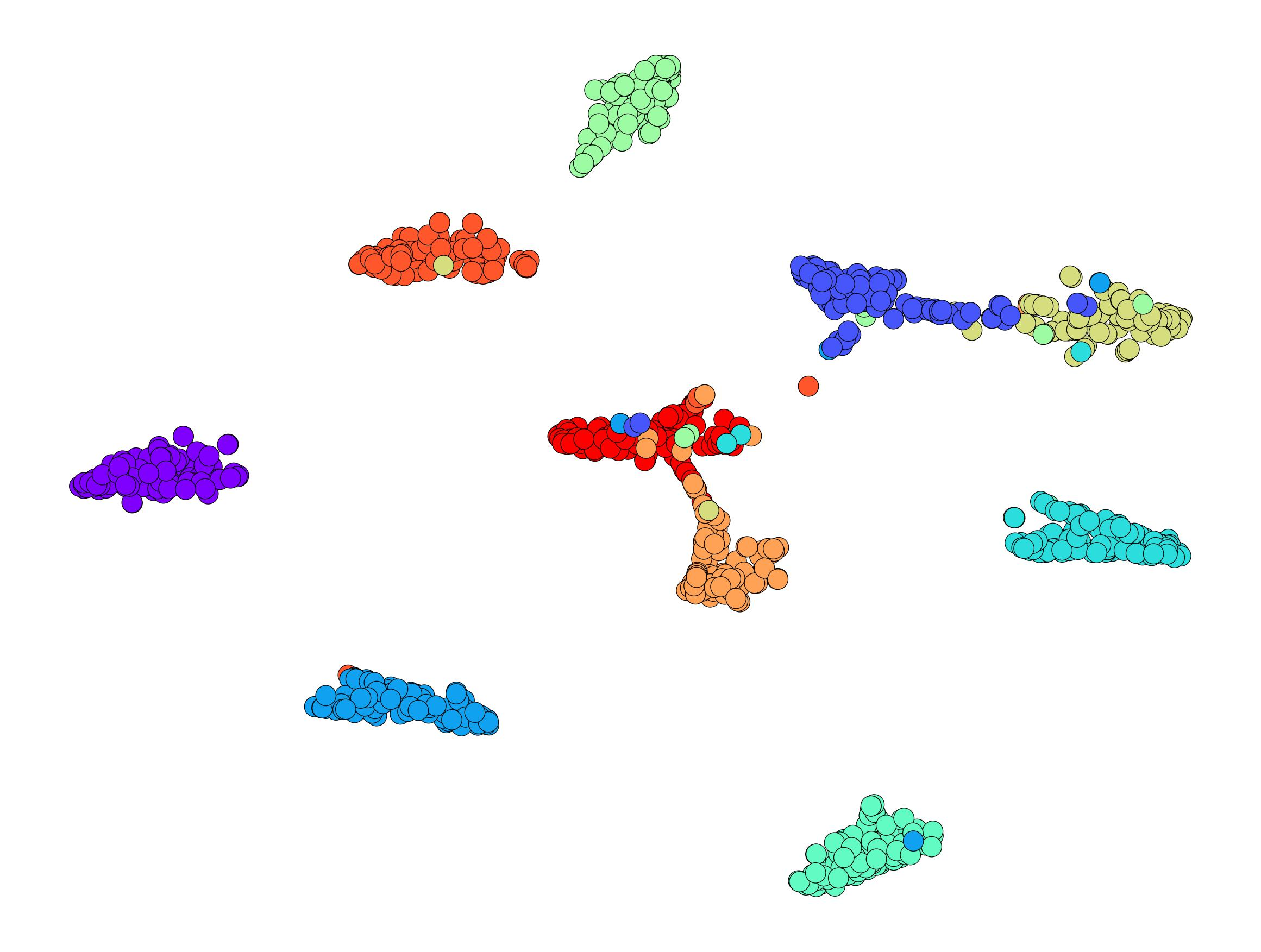}} 
\hspace{3mm}
\subfigure[MM-GANN-DDI]{\includegraphics[width=1.6in,height=1.1in]{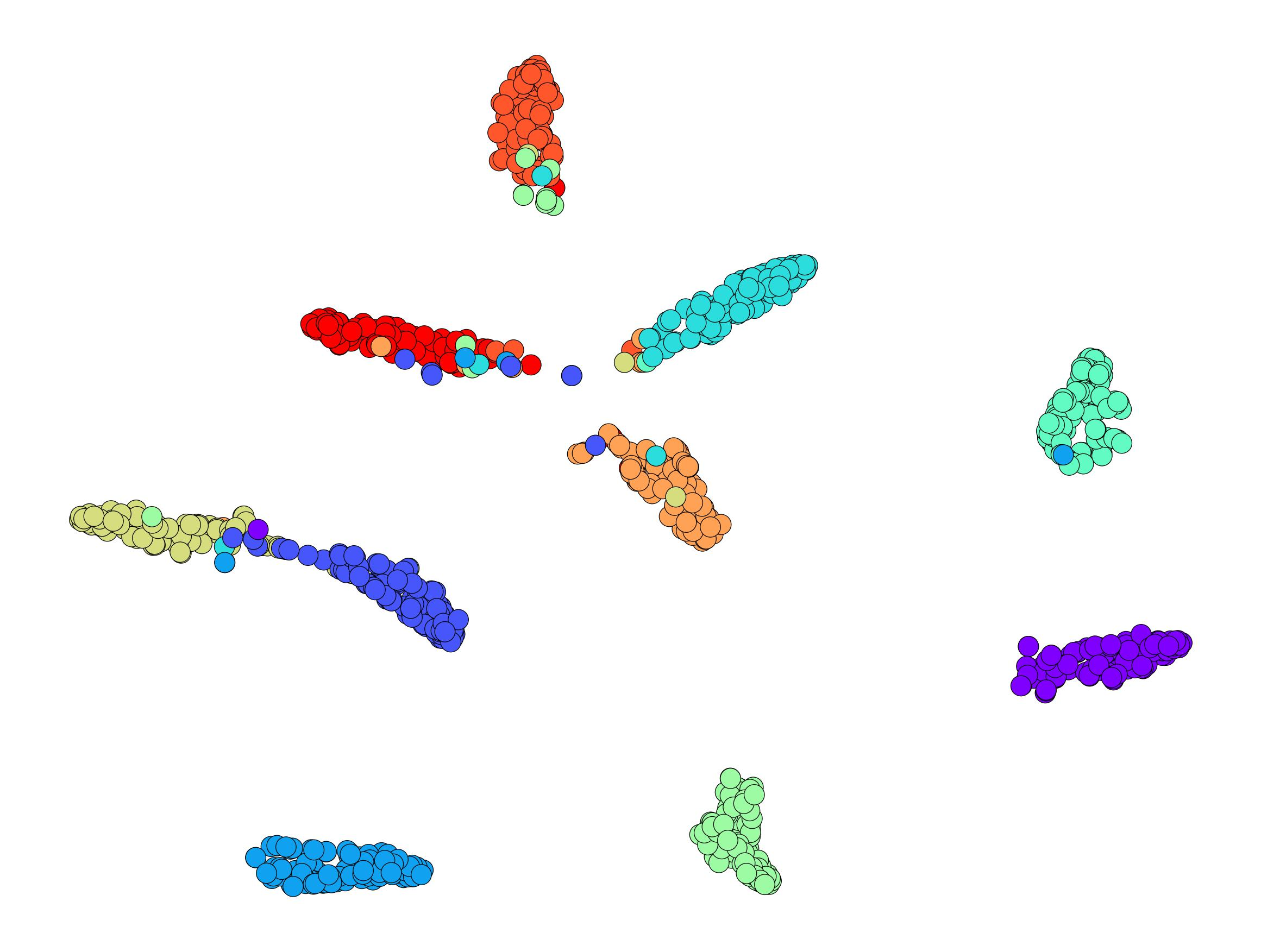}} 
\hspace{3mm}
\subfigure[HMGRL]{\includegraphics[width=1.6in,height=1.1in]{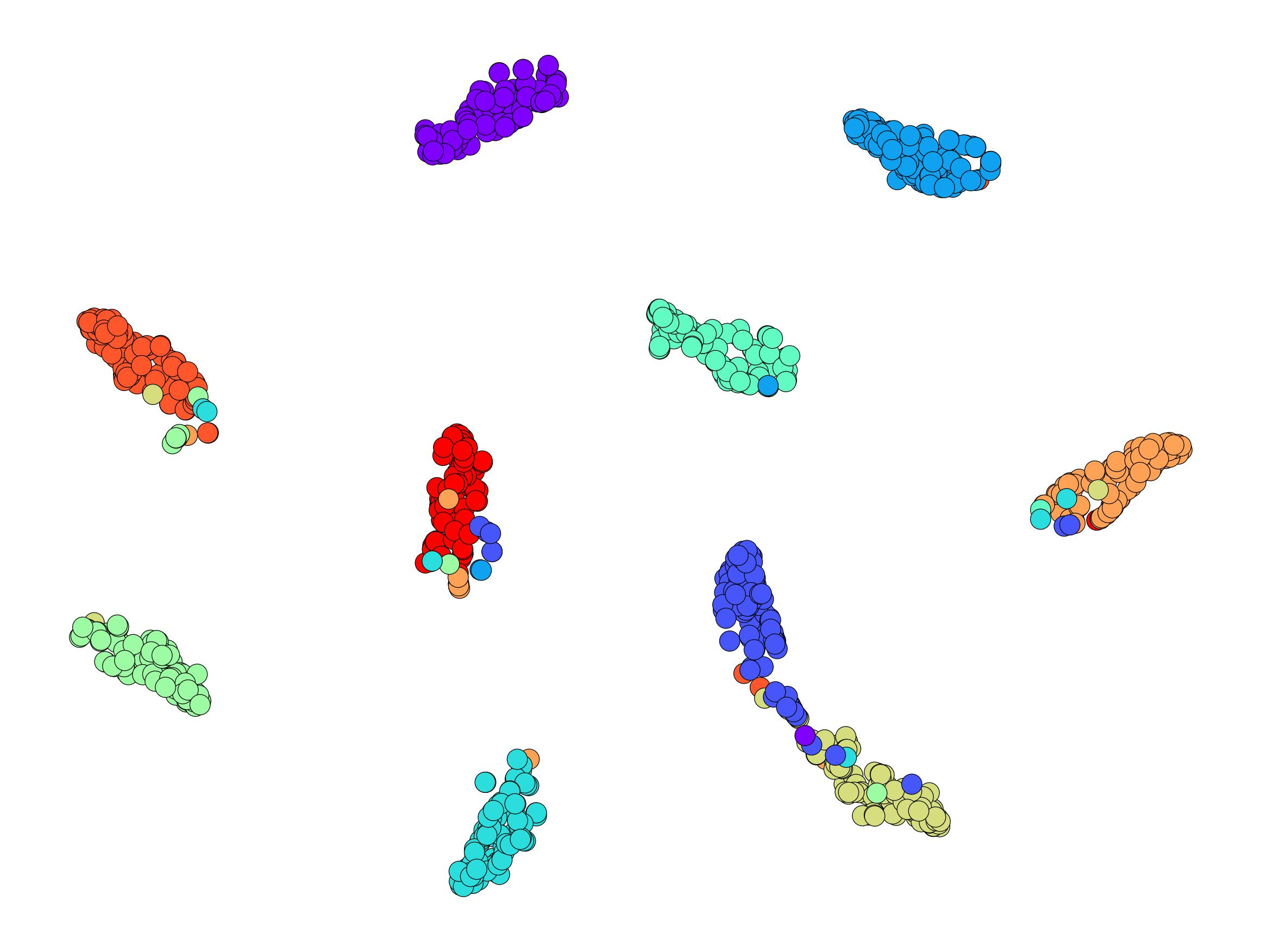}} 
\caption{2D visualization obtained by raw data and different prediction methods on Task 1 of Dataset 1. Points represent drug pairs, and the colors denote different interaction event types.}
\label{points}
\vspace{-0.3cm}
\end{figure*}

\vspace{-0.2cm}
\subsection{Comparison With Other Methods}\label{Comparison}
The robustness of our HMGRL is evaluated by comparing it with several SOTA DDI prediction methods.
These methods include DDIMDL, RANEDDI, MDF-SA-DDI, MAEDDIE, MCFF-MTDDI, and MM-GANN-DDI.
The results are summarized in Table~\ref{dataset1}, encompassing three different tasks from Dataset 1.
In most cases, our HMGRL outperforms the competitors across all three tasks. This comparison demonstrates the effectiveness of our HMGRL in DDI prediction.
In Task 1, although RANEDDI also learns the drug RaGSEs, HMGRL outperforms it. 
The main reason causing this phenomenon is that RANEDDI ignores many valuable biological attributes, resulting in limited generalization ability for RANEDDI.
MDF-SA-DDI and MAEDDIE consider various drug attributes and use multiple drug fusion methods to learn DP features. Even so, our HMGRL still performs better.
The primary advantage of HMGRL stems from MDF-SA-DDI and MAEDDIE's failure to harness labeled drug interaction information.

To gain a deeper understanding, we compared the performance of our HMGRL with six baseline methods across different events. Fig.~\ref{zzzz} showcases both AUPR and AUC scores for these methods on each event within Task 1 of Dataset 1.
The results show that HMGRL consistently outperforms the baselines, delivering superior AUPR and AUC scores for most event types. This demonstrates the robustness of HMGRL in handling diverse event types.
However, it's worth noting that for certain low-frequency events, specifically $\#39$, $\#52$, and $\#64$, HMGRL's performance somewhat diminishes. These events have limited training samples, with frequencies of $98$, $24$, and $10$, respectively. The constrained sample size for these events might be the primary factor leading to the suboptimal performance observed.

To demonstrate the superiority of HMGRL visually, we employ t-SNE \cite{t-SNE} to visualize the distribution of DP representations produced by raw data and different DDI prediction methods. We select 100 DPs from each of the DDI event types numbered from $\#1$ to $\#10$ within Task 1 of Dataset 1. As highlighted in Fig. \ref{points}, the visualized results underscore that HMGRL possesses a more distinct structure, aptly capturing the intrinsic clustering structure in the data.

Tasks 2 and 3 benchmark our HMGRL against five competitive prediction methods.
We exclude RANEDDI from this comparison as it is specifically designed for representation learning of known drugs.
The test DDIs in Tasks 2 and 3 involve new drugs. Thus, RANEDDI does not apply to Tasks 2 and 3.
Moreover, the absence of interaction information for new drugs weakens the generalization ability of prediction methods on Tasks 2 and 3.
As a result, the metrics of all DDI prediction methods in Tasks 2 and 3 tend to be lower than those observed in Task 1.
Despite these challenges, HMGRL outshines the competition in most cases.
MM-GANN-DDI also integrates multiple biological attributes and a multi-relational DDI graph to learn drug embeddings, while our HMGRL still surpasses it. The enhancement in performance can be ascribed to various factors. Firstly, our HMGRL considers implicit correlations between DPs. Furthermore, our HMGRL learns effective RaGSEs for new drugs.

Table \ref{dataset222} details six metrics obtained from various methods across three tasks of Dataset 2. 
Compared with Dataset 1, Dataset 2 encompasses a broader spectrum of DDI event information. Thus, Dataset 2 is an ideal benchmark for evaluating the model's performance on a more heterogeneous large-scale dataset.
Notably, our HMGRL surpasses the competing approaches in most cases, excelling across all performance metrics over the three tasks. This comparative analysis underscores the effectiveness of HMGRL, verifying its capability for accurate and robust large-scale prediction of DDIs.
The superiority of HMGRL can be attributed to three primary factors:
First, HMGRL takes full advantage of the explicit physical connections between drugs, including similarities and interaction relationships.
Second, HMGRL learns DP features using multiple sources, improving the model's generalization.
Third, HMGRL captures underlying associations between DPs from multiple perspectives.

\subsection{Case Study}\label{Case}
\begin{table}[t] \scriptsize% \\tiny\\scriptsize\\footnotesize\\small\\normalsize\\large\\Large\\LARGE\\huge
\caption{The drug names and event types of the confirmed DDIs.}
\vspace{-0.2cm}
\tabcolsep=1pt%%
\begin{tabular}{lllll}
\toprule
Rank  & Drug names  & DDI event type\\
\midrule
1 & Fedratinib and Quercetin  & The metabolism decrease\\
2 & Doravirine and Venetoclax & The metabolism decrease\\
\multirow{2}{*}{$3$} & \multirow{2}{*}{Bleomycin and Ribavirin} &The excretion rate which could result in a \\
         &   & higher serum level decrease\\  
4 & Teriflunomide  and Fluocinonide & The risk or severity of adverse effects increase\\
5 & Mirabegron and Doxylamine & The risk or severity of adverse effects increase\\
6 & Artesunate and Cocaine & The risk or severity of adverse effects increase\\
7 & Maprotiline and Paliperidone  & The risk or severity of adverse effects increase\\
8 & Dexamethasone and Zolpidem  & The metabolism increase\\ 
% 9 & Prednisolonephosphate and Tramadol & The metabolism increase\\
9 & Hydrocodone and Betamethasone & The metabolism increase\\
10 & Zopiclone and Betamethasone & The metabolism increase\\
11 & Benidipine and Fluoxetine  & The risk or severity of QTc prolongation increase\\ 
12 & Mirabegron and Macimorelin & The risk or severity of QTc prolongation increase\\
13 & Sunitinib and Granisetron & The risk or severity of QTc prolongation increase\\
14 & Sunitinib and Dolasetron & The risk or severity of QTc prolongation increase\\
\bottomrule
\label{casestudy}
\vspace{-0.4cm}
\end{tabular}
\end{table}

In this section, we conduct case studies to demonstrate the efficacy of our HMGRL approach. 
To this end, we employ the complete set of DDIs along with their corresponding event types from Dataset 2 as the training data for our HMGRL.
Once trained, we leverage the model to assess the remaining DPs.
Considering a set of 1,000 drugs, there are 293,471 remaining DPs after excluding 206,029 pairs known to interact. 
Drugbank is a public drug database updated regularly to reflect new scientific research and medical information.
Therefore, it is likely that numerous unknown interactions exist among these 293,471 remaining DPs. 
We pay attention to five events with the highest frequencies and check up on the top 20 predictions related to each event.
Furthermore, we test the 100 DPs using the DDI Checker tool\footnote{\url{https://go.drugbank.com/interax/multi_search}}.
The test results reveal that 14 out of the 100 evaluated DDIs are successfully predicted.
This result proves that our HMGRL performs well on large-scale data.
Meanwhile, the capability of our HMGRL to discover novel DDIs is also confirmed.
Table \ref{casestudy} documents these 14 correct predictions.
As an illustration, the metabolism of Doravirine can be decreased when combined with Venetoclax. The risk or severity of QTc prolongation and torsade de pointes can be increased when Sunitinib is combined with Dolasetron. These findings exemplify the precision and practical applicability of our HMGRL in predicting complex drug interactions.

\vspace{-0.3cm}
\section{Conclusion}\label{Conclusion}
Our HMGRL captures diverse explicit relationships among drugs and multifaceted implicit associations between DPs to improve representation learning for DPs.
In implementing HMGRL, we start with drug knowledge discovery to construct drug interaction graphs with clear physical connections.
Then, we employ the RGCN to learn features from these drug interaction graphs, capturing diverse explicit relationships among drugs.
Additionally, leveraging SC's capability to craft graphs from data and discern intricate structures, we innovatively craft an MVDSC module.
This advanced MVDSC effectively identifies complex data structures and is additionally optimized for downstream tasks.
Through MVDSC, our HMGRL adeptly identifies underlying correlations among DPs from multiple perspectives.
Overall, our work provides a novel and promising solution for DP representation learning.
Furthermore, rich drug information and a well-designed network structure in HMGRL enable accurate large-scale DDI prediction. 
Our future work will consider more drug-related knowledge to improve the current HMGRL scheme for enhancing the inferential capabilities of the network further.

\begin{table}[t]\footnotesize % \\tiny\\scriptsize\\footnotesize\\small\\normalsize\\large\\Large\\LARGE\\huge
\caption{The hyperparameters of best accuracy for the proposed HMGRL on all Tasks.}
\vspace{-0.2cm}
\tabcolsep=2pt%%
\begin{tabular}{lllllllllllll}
\toprule
Dataset & Task & $bs$ & $lr$& $dr$& $te$& $d'$& $L$ & $d^{att}$& $d^{emb}$& $M$& $C$ & $\alpha$  \\
\midrule
\multirow{3}{*}{Dataset$~$1 }
& Task 1   & 512 & 2e-5 & 0.3 &120& 500 &0& 1500 & 200 & 5 & 200 & 0.2\\
& Task 2   & 1024& 5e-6 & 0.2 &120& 500 &3& 800  & 800 & 5 & 400 & 0.5\\
& Task 3   & 1024& 5e-6 & 0.3 &120& 500 &3& 800  & 800 & 5 & 400 & 0.5\\
\midrule
\multirow{3}{*}{Dataset$~$2 }
&Task 1   &1024 &2e-5 &0.3 &150 &500 &0& 1500 & 200 &5 & 400 &0.2\\
&Task 2   &1024 &5e-6 &0.4 &150 &500 &3& 800  & 800 &5 & 400 &0.5\\
&Task 3   &1024 &5e-6 &0.4 &150 &500 &3& 800  & 800 &5 & 400 &0.5\\ 
\bottomrule
\label{parasearch}
\end{tabular}
\vspace{-0.6cm}
\end{table}

\bibliographystyle{IEEEbib}
\bibliography{refs}
 
\begin{IEEEbiography}
[{\includegraphics[width=1in,height=1.25in]{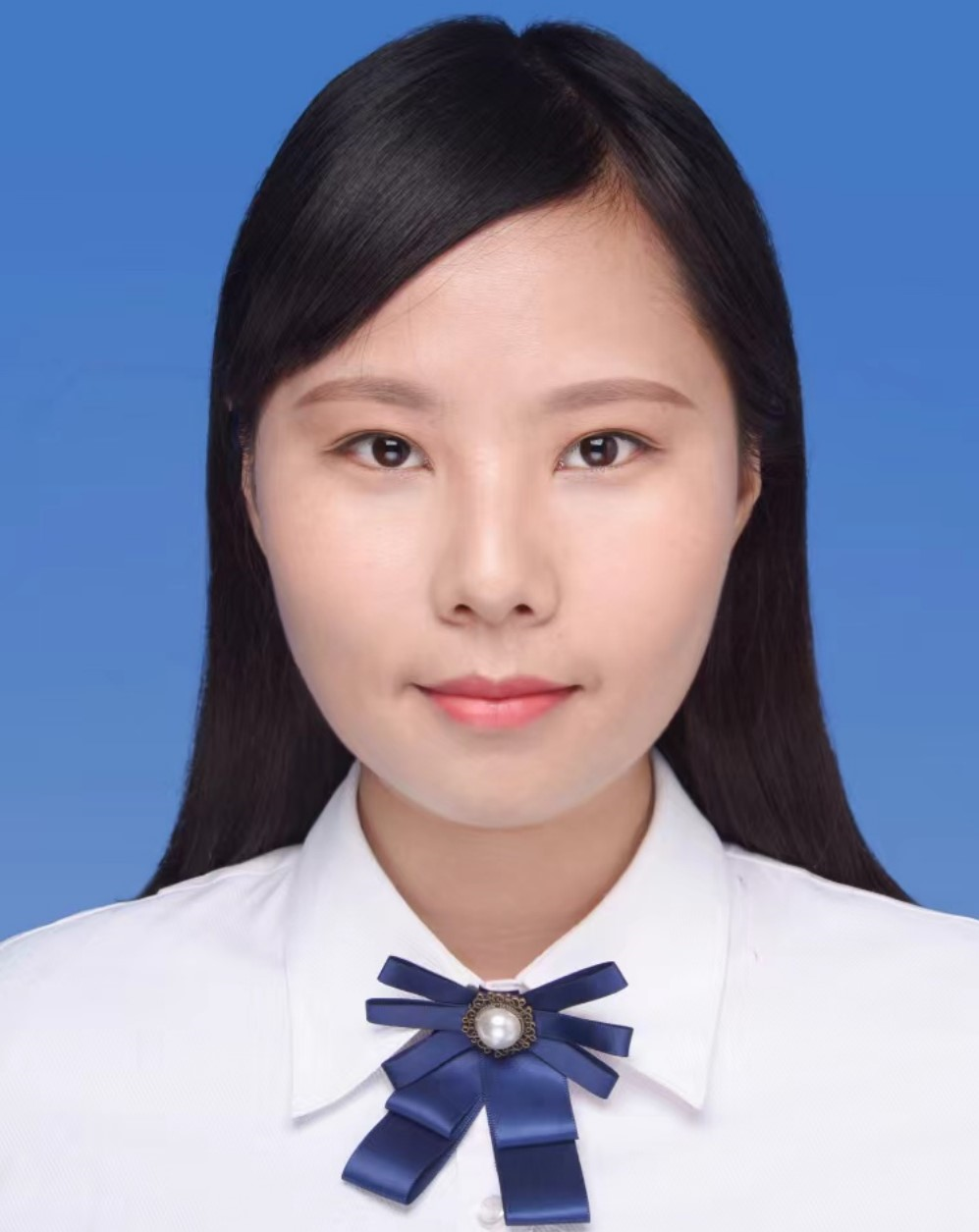}}]{Mengying Jiang} received her M.S. degree from the Department of Communication Engineering, Guangdong University of Technology, Guangzhou, in 2019. She is a Ph.D. candidate in Electronic and Information Engineering at Xi’an Jiaotong University, Xi’an, China.  Her research interests include machine learning, graph neural networks, representation learning, hyperspectral image classification, reinforcement learning, and molecular interaction prediction.
\end{IEEEbiography}
\vspace{-2cm}
\begin{IEEEbiography}
[{\includegraphics[width=1in,height=1.25in]{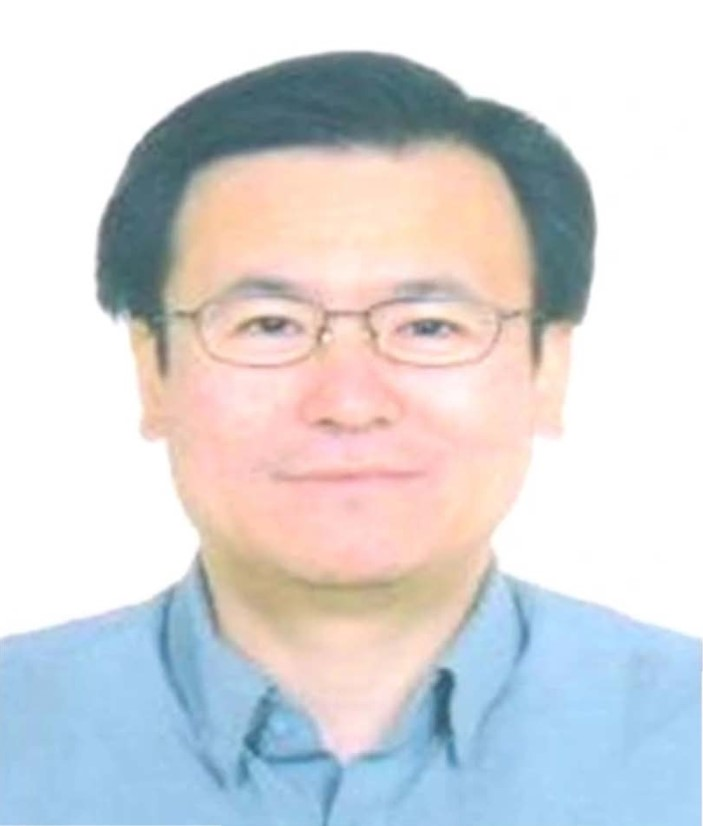}}]{Guizhong Liu} (Member, IEEE) received the B.S. and M.S. degrees in computational mathematics from Xi’an Jiaotong University, Xi’an, China, in 1982 and 1985, respectively, and the Ph.D. degree in mathematics and computing science from the Eindhoven University of Technology, Eindhoven, The Netherlands, in 1989., He is a Full Professor at the School of Electronics and Information Engineering at Xi’an Jiaotong University. His research interests include nonstationary signal analysis and processing, image processing, audio and video compression, inversion problems, graph neural networks, reinforcement learning, and molecular interaction prediction.
\end{IEEEbiography}
\vspace{-2cm}
\begin{IEEEbiography}
[{\includegraphics[width=1in,height=1.25in]{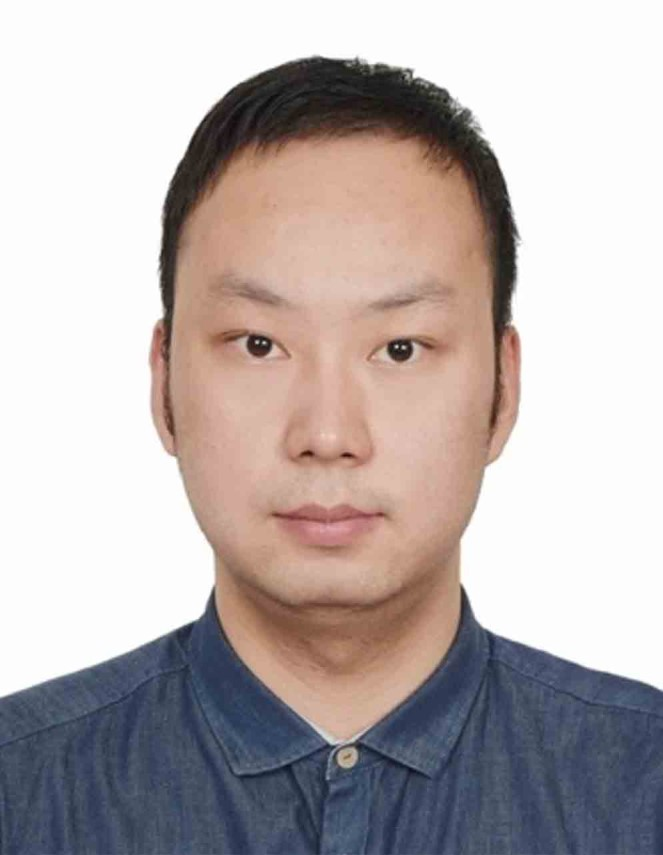}}]{Yuanchao Su} (Senior Member, IEEE) received the B.S. and M.Sc. degrees from the Xi’an University of Science and Technology, Xi’an, China, in 2012 and 2015, respectively, and the Ph.D. degree from Sun Yat-sen University, Guangzhou, China, in 2019., From 2013 to 2015, he was an Exchange Postgraduate with the Optical Laboratory, Institute of Remote Sensing and Digital Earth, Chinese Academy of Sciences, Beijing, China. From 2018 to 2019, he was a Visiting Researcher with the Advanced Imaging and Collaborative Information Processing Group, Department of Electrical Engineering and Computer Science, University of Tennessee, Knoxville, TN, USA. In 2019, he joined the Department of Remote Sensing, College of Geomatics, Xi’an University of Science and Technology, where he is currently an Associate Professor. His research interests include remote sensing image processing, hyperspectral unmixing, hyperspectral image classification, neural networks, and deep learning.
\end{IEEEbiography}
\vspace{-2cm}
\begin{IEEEbiography}
[{\includegraphics[width=1in,height=1.25in]{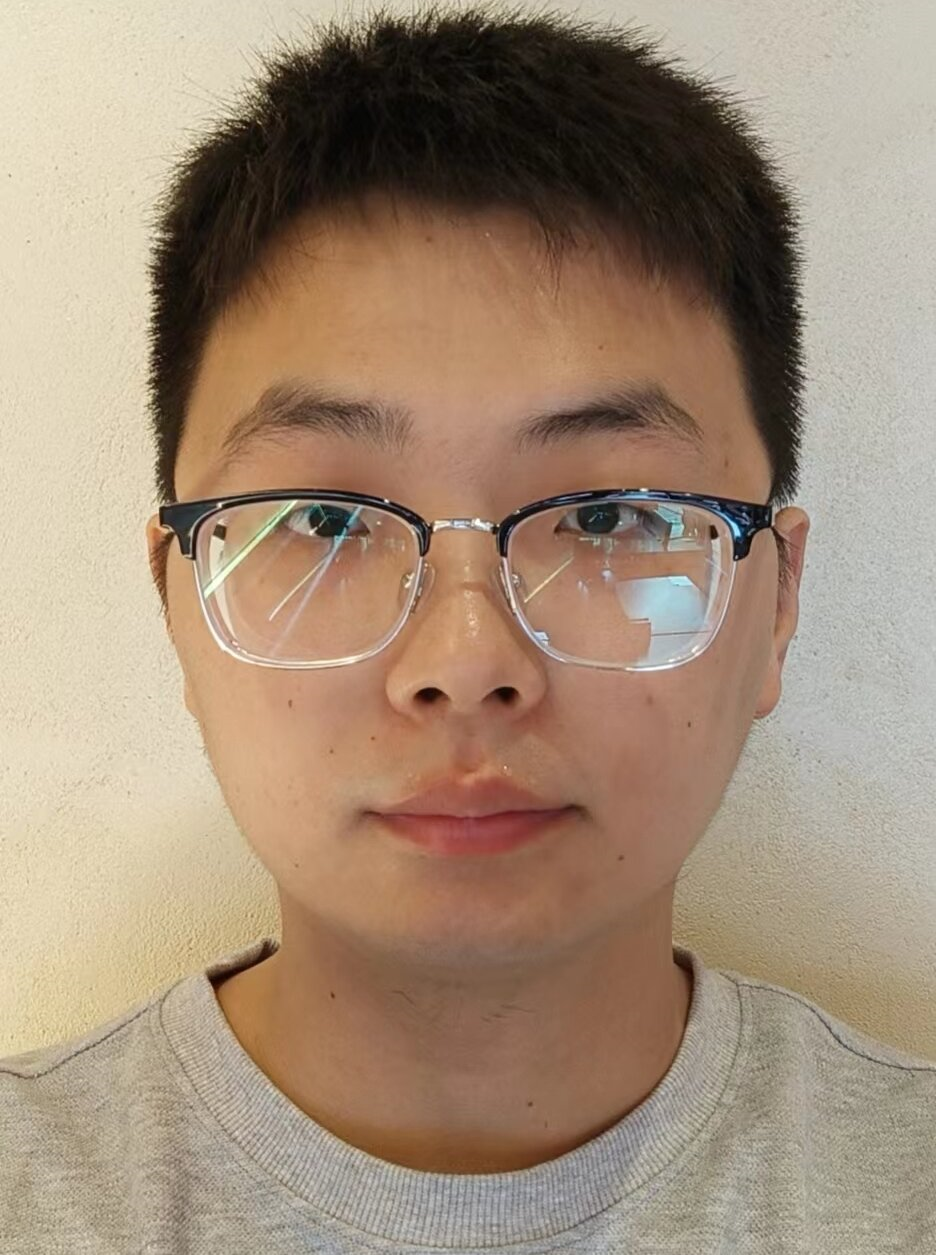}}]{Weiqiang Jin} received his M.S. degree in Computer Science from Shanghai University, Shanghai, in 2022. He is a Ph.D. candidate in Electronic and Information Engineering at Xi'an Jiaotong University, Xi'an, China.
His main research interests are 1) natural language processing, including knowledge graph-related question answering, aspect-term-based Sentiment analysis, information retrieval, and prompt learning. 2) Reinforcement Learning, especially in researching the multi-agent information interaction mechanism. 
\end{IEEEbiography}
\vspace{-2cm}
\begin{IEEEbiography}
[{\includegraphics[width=1in,height=1.35in]{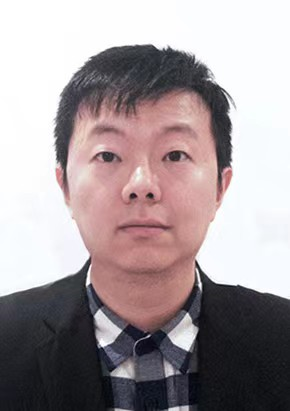}}]{Biao Zhao} obtained his B.S. degree in information engineering from Xi’an Jiaotong University in 2009. He then pursued graduate studies at Politecnico di Torino, obtaining his Master of Science in Communication Engineering in 2011. Zhao continued at Politecnico di Torino for doctoral studies, obtaining his Ph.D. in Electronics and Communication Engineering in 2015.
He has joined Xi'an Jiaotong University since 2019.
He is currently an associate professor at Xi'an Jiaotong University, specializing in reinforcement learning, computer vision, pattern recognition, machine learning, etc.

\end{IEEEbiography}
 
\end{document}